\documentclass[11pt]{article}

\usepackage[preprint]{acl}

\usepackage{times}
\usepackage{latexsym}

\usepackage[T1]{fontenc}

\usepackage[utf8]{inputenc}

\usepackage{microtype}

\usepackage{inconsolata}

\usepackage{graphicx}

\usepackage{booktabs}
\usepackage{multirow}
\usepackage[table,xcdraw]{xcolor}

\title{MMReview: A Multidisciplinary and Multimodal Benchmark for LLM-Based Peer Review Automation}

\author{
 \textbf{Xian Gao\textsuperscript{1}},
 \textbf{Jiacheng Ruan\textsuperscript{1}},
 \textbf{Zongyun Zhang\textsuperscript{1}},
 \textbf{Jingsheng Gao\textsuperscript{1}},
\textbf{Ting Liu\textsuperscript{1}},
\textbf{Yuzhuo Fu\textsuperscript{1}},
\\
\\
 \textsuperscript{1}Shanghai Jiao Tong University,
\\
 \small{
   gaoxian@sjtu.edu.cn
 }
}

\begin{document}
\maketitle
\begin{abstract}

With the rapid growth of academic publications, peer review has become an essential yet time-consuming responsibility within the research community. Large Language Models (LLMs) have increasingly been adopted to assist in the generation of review comments; however, current LLM-based review tasks lack a unified evaluation benchmark to rigorously assess the models’ ability to produce comprehensive, accurate, and human-aligned assessments, particularly in scenarios involving multimodal content such as figures and tables. To address this gap, we propose \textbf{MMReview}, a comprehensive benchmark that spans multiple disciplines and modalities. MMReview includes multimodal content and expert-written review comments for 240 papers across 17 research domains within four major academic disciplines: Artificial Intelligence, Natural Sciences, Engineering Sciences, and Social Sciences. We design a total of 13 tasks grouped into four core categories, aimed at evaluating the performance of LLMs and Multimodal LLMs (MLLMs) in step-wise review generation, outcome formulation, alignment with human preferences, and robustness to adversarial input manipulation. Extensive experiments conducted on 18 open-source models and 3 advanced closed-source models demonstrate the thoroughness of the benchmark. We envision MMReview as a critical step toward establishing a standardized foundation for the development of automated peer review systems.

\end{abstract}
\section{Introduction}

Peer review is essential to scholarly publishing, ensuring research quality and enhancing academic writing. However, the growing volume of submissions has strained the traditional review process, leading to inefficiencies and limited reviewer availability~\cite{kimPositionAIConference2025a}, which restricts feedback and delays review outcomes. Advances in LLMs have made automated peer review increasingly viable, as these models show strong reasoning abilities and can offer constructive feedback on academic manuscripts~\cite{liuReviewerGPTExploratoryStudy2023,zhaoWordsWorthNewborn2024,zhuangLargeLanguageModels2025}, partially alleviating reviewer burden. Yet, current evaluations of LLM-generated reviews focus mainly on final outputs, lacking in-depth analysis of the reasoning processes behind model judgments. Additionally, most studies concentrate on AI papers with publicly available text, overlooking the multimodal nature of academic papers, such as figures and tables, and the evaluation of LLMs in reviewing research across broader scientific domains.

To address the aforementioned challenges, we propose \textbf{MMReview}, a comprehensive benchmark for peer review generation that spans multiple disciplines and modalities. MMReview incorporates three distinct types of input modalities: textual content from manuscripts, figures and tables embedded within the papers, and rendered PDF pages converted into images. These data span 17 research domains across 4 disciplinary categories. To obtain high-quality peer review samples for evaluation purposes, we developed a multi-model collaborative pipeline for data filtering and generation. Specifically, we first curated a total of 51,881 papers with associated reviews; then, we filtered the collected seed dataset D to obtain high-quality papers while maintaining a relatively balanced distribution; subsequently, we extracted reference answers from human reviews. Finally, we conducted manual verification to correct errors, resulting in a curated set of 240 samples that serve as the foundation for task construction. Building upon these samples, we introduce 4 thematic categories encompassing 13 diverse tasks, each designed to thoroughly assess the capabilities of LLMs in step-wise review generation, outcome formulation, alignment with human preferences, and robustness to adversarial input manipulation. 
We conduct comprehensive experiments on 16 open-source models as well as 5 state-of-the-art closed-source models, including GPT-4o and Claude-4-Sonnet, across 13 tasks. The results highlight the comprehensive nature of the MMReview benchmark and uncover several key findings, offering insights for future research on LLM-based automated academic peer review.

The primary contributions of this paper can be summarized as follows:

\begin{itemize}
    \item We introduce \textbf{MMReview}, the first comprehensive evaluation benchmark for automated academic peer review using LLMs, spanning multiple disciplines and modalities. Built upon our data filtering and generation pipeline, MMReview comprises 240 high-quality samples across 17 academic fields in 4 disciplines.

    \item We meticulously design 13 distinct tasks encompassing a total of 6,724 thoughtfully curated questions, enabling multi-dimensional evaluation of model performance. These diverse tasks allow for targeted assessment and facilitate the identification of potential limitations in LLM-generated peer review content.

    \item We conduct extensive experiments on 18 open-source and 3 closed-source models using the MMReview benchmark, offering some key findings of LLM-based automated reviewing. Our findings offer in-depth analysis and valuable guidance for the future development of LLM-assisted peer review systems.
\end{itemize}

\section{Related Works}
\subsection{LLMs for Paper Review}

LLMs have shown strong potential in analyzing complex scholarly texts~\cite{liuReviewerGPTExploratoryStudy2023,zhaoWordsWorthNewborn2024,zhuangLargeLanguageModels2025}. Initial studies indicate that LLM-generated review comments partially overlap with those of human reviewers, suggesting their potential contribution to peer review~\cite{robertsonGPT4SlightlyHelpful2023,liangCanLargeLanguage2023}. However, further research reveals that even advanced models like GPT-4o often fail to meet human expectations in review quality~\cite{zhouLLMReliableReviewer2024}. To improve alignment with peer review standards, researchers have built datasets from public review platforms and fine-tuned LLMs~\cite{kangDatasetPeerReviews2018,yuanCanWeAutomate2021,shenMReDMetaReviewDataset2022,dyckeNLPeerUnifiedResource2023,gaoReviewer2OptimizingReview2024}. Other methods involve multi-turn dialogue~\cite{tanPeerReviewMultiTurn2024} or multi-agent prompting~\cite{darcyMARGMultiAgentReview2024} to generate comprehensive feedback. Nonetheless, these efforts focus solely on textual reviews, neglecting the vital role of multimodal content, such as figures and tables, and lack rigorous analysis of the reasoning processes behind LLM-generated critiques.

\subsection{Evaluation for LLM-based Peer Review}

Prior studies~\cite{shenMReDMetaReviewDataset2022,yuAutomatedPeerReviewing2024,gaoReviewer2OptimizingReview2024,tanPeerReviewMultiTurn2024,gaoReviewAgentsBridgingGap2025} have predominantly evaluated the quality of LLM-generated peer review comments by measuring their correlation or similarity with human-written reviews using automated metrics such as BLEU~\cite{papineniBleuMethodAutomatic2002}, ROUGE~\cite{linROUGEPackageAutomatic2004}, BERTScore~\cite{zhangBERTScoreEvaluatingText2020}, and METEOR~\cite{banerjeeMETEORAutomaticMetric2005}. In addition, several studies~\cite{robertsonGPT4SlightlyHelpful2023,zhouLLMReliableReviewer2024,gaoReviewAgentsBridgingGap2025} have adopted the \textit{LLM-as-a-judge} paradigm, leveraging cutting-edge language models to assess the quality of review comments produced by other LLMs. Given the absence of an established gold standard for this evaluation task, recent research~\cite{xuBenchmarkingLLMsJudgments2024a} has introduced the Generative Estimator for Mutual Information (GEM) to quantify the degree of semantic overlap between LLM-generated and human-authored reviews. Nevertheless, existing evaluation methodologies are not grounded in a unified benchmark or task framework, and they fall short of providing a comprehensive analysis of the underlying reasoning processes involved in LLM-generated peer review.

\begin{figure*}
    \centering
    \includegraphics[width=\linewidth]{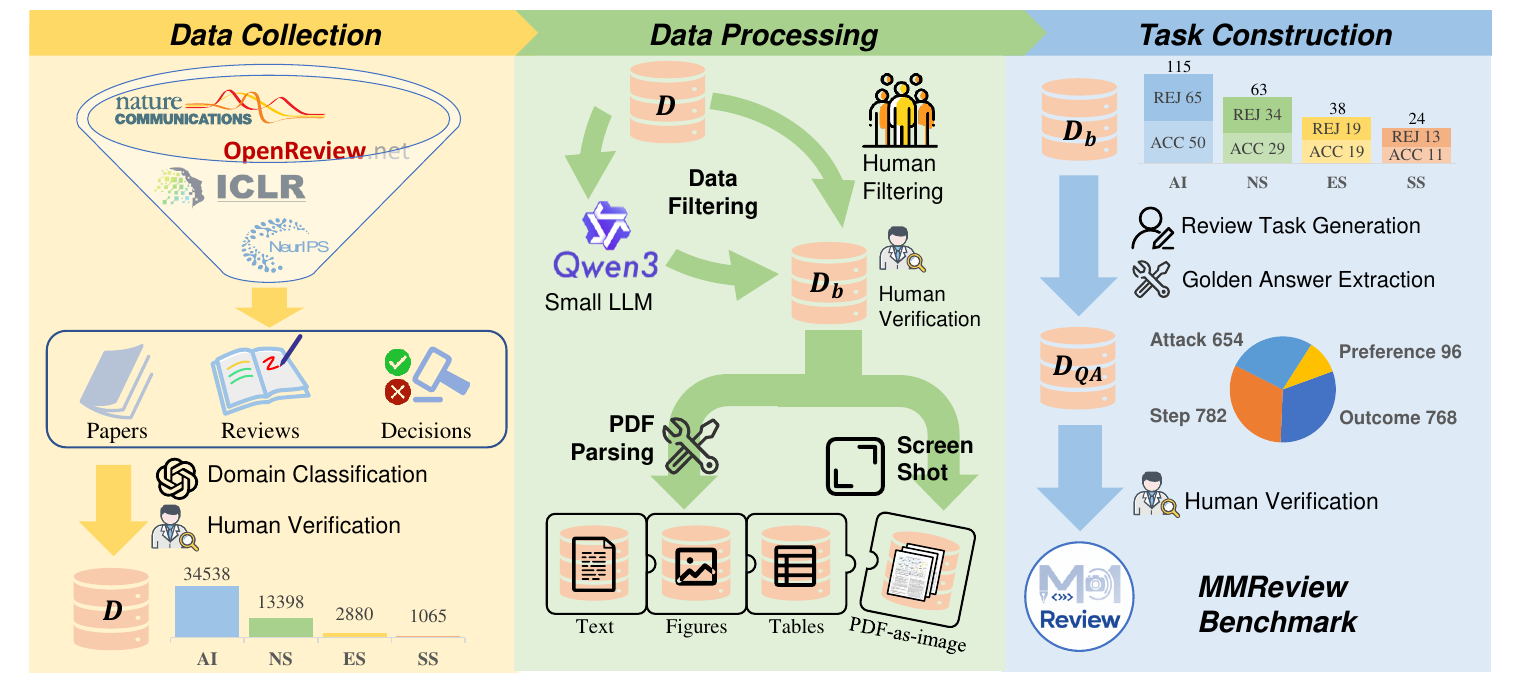}
    \caption{The construction pipeline of MMReview. The construction pipeline is divided into three stages: data collection, data processing, and task construction.}
    \label{fig:pipeline}
\end{figure*}

\section{MMReview Benchmark}

In this section, we first present the overall pipeline for data collection and construction of the MMReview benchmark, followed by a detailed exposition of the task design methodology.

\subsection{Overall Pipeline of MMReview Benchmark}

As illustrated in Figure \ref{fig:pipeline}, the construction of MMReview is divided into three stages: data collection, data processing, and task construction.

\subsubsection{Data Collection}

During the data collection phase, we gathered academic papers from publicly accessible peer review platforms or sources where reviewer comments were openly available. These papers not only contain the full manuscript texts but also include reviewer-written evaluations and final decisions (accept or reject). Specifically, we curated a total of 51,881 papers with associated reviews submitted between 2013 and 2024 to venues such as ICLR and NeurIPS (from the OpenReview platform
and NeurIPS Proceedings
, as well as articles from the journal Nature Communications
. For each collected paper, we employed Deepseek-V3 and GPT-4o to automatically infer its academic discipline and research domain. In cases where the two models produced inconsistent classifications, human verification was performed. Ultimately, all papers were categorized into four overarching disciplines: Artificial Intelligence, Natural Sciences, Engineering Sciences, and Social Sciences, resulting in a seed dataset denoted as $D$. Figure~\ref{fig:pipeline} illustrates the distribution of papers across the four disciplines within $D$.

\begin{table}[!t]
\centering
\small
\setlength{\tabcolsep}{2.0pt}
\resizebox{\columnwidth}{!}{%
\begin{tabular}{@{}cccc@{}}
\toprule
\textbf{Discipline}                                                              & \textbf{Research Field}       & \textbf{\# Samples} & \textbf{ACC:REJ} \\ \midrule
\multirow{7}{*}{\begin{tabular}[c]{@{}c@{}}Artificial\\ Intelligence\end{tabular}} & Machine Learning              & 17                  & 8:9              \\
 & Computer Vision                 & 20 & 12:8 \\
 & Natural Language Processing     & 16 & 7:9  \\
 & Reinforcement Learning          & 11 & 3:8  \\
 & Graph Neural Networks           & 19 & 8:11 \\
 & Signal Processing               & 17 & 8:9  \\
 & AI Application                  & 15 & 4:11 \\ \midrule
\multirow{5}{*}{\begin{tabular}[c]{@{}c@{}}Natural\\ Sciences\end{tabular}}        & Biology \& Medicine           & 11                  & 5:6              \\
 & Physics                         & 16 & 7:9  \\
 & Chemistry                       & 13 & 5:8  \\
 & Environmental \& Earth Sciences & 10 & 7:3  \\
 & Mathematics \& Statistics       & 13 & 5:8  \\ \midrule
\multirow{4}{*}{\begin{tabular}[c]{@{}c@{}}Engineering\\ Sciences\end{tabular}}    & Materials Science             & 14                  & 6:8              \\
 & Control Science                 & 12 & 5:7  \\
 & Electronic Information          & 6  & 4:2  \\
 & Energy Science                  & 6  & 4:2  \\ \midrule
\begin{tabular}[c]{@{}c@{}}Social\\ Sciences\end{tabular}                          & Society, Economics \& Finance & 24                  & 11:13            \\ \bottomrule
\end{tabular}%
}
\caption{The distribution of papers across various research domains.}
\label{tab:domain}
\end{table}

\subsubsection{Data Processing}

During the data processing stage, we first filtered the collected seed dataset $D$ to obtain high-quality samples while maintaining a relatively balanced distribution across disciplines and ensuring a comparable number of accepted and rejected papers. To achieve this, we designed a dual-path joint data selection mechanism that simultaneously filters for sample quality and enforces distributional balance.

Specifically, we utilized Qwen3-32B\footnote{Qwen3-32B ranks among the top 10 on the OpenCompass leaderboard, offering a favorable trade-off between performance and model size, making it efficient for large-scale sample filtering}~\cite{yang2025qwen3technicalreport} to generate summaries of each paper under two distinct input conditions: one using only the abstract and the other using the full text of the manuscript. A greater divergence between the two generated summaries is interpreted as evidence that the full text provides substantially more information, thus indicating higher sample quality. Samples with a significant information gain from the full text were retained in our test benchmark. To further ensure that the benchmark maintains a balanced distribution across academic disciplines and an approximately equal ratio of accepted to rejected papers, we supplemented the benchmark by manually incorporating top-ranked papers from specific domains in $D$ based on the quality rankings. This human-filtering procedure guarantees that the composition of the benchmark dataset aligns with our desired distributional properties.

During the data filtering phase, we obtained a total of 240 paper samples spanning 17 research domains across 4 major disciplines to construct our evaluation benchmark, denoted as $D_b$. The statistical details of $D_b$, including the number of samples per domain and the distribution of accepted versus rejected papers, are presented in Table~\ref{tab:domain}. For each discipline, we ensured a relatively balanced number of samples across research fields while approaching the actual acceptance/rejection ratio through a combination of model-based filtering and manual curation. For each of these 240 samples, we utilized PDF parsing tools to extract textual content, figures, and tables from the manuscript files, and converted each page of the PDF into corresponding images. As a result, we constructed three distinct modalities of input data: \textit{text-only}, \textit{multimodal} (text combined with extracted visual elements), and \textit{PDF-as-image}.

\subsubsection{Task Construction}

During task construction, we developed 13 tasks grouped into four thematic categories, \textit{step-based}, \textit{outcome-based}, \textit{preference-based}, and \textit{attack-based}, reflecting the peer review workflow and challenges LLMs may face. Task distribution is shown in Figure~\ref{fig:pipeline}. Prompts were designed based on reviewer guidelines from major academic conferences. For each task, we used regular expressions or GPT-4o to extract reference answers from human reviews in $D_b$, forming the question–answer dataset $D_{QA}$. For the \textit{Fake Strength Evaluation} and \textit{Fake Weakness Evaluation} tasks, GPT-4o and custom prompts generated antonymic rewrites, which were manually verified for semantic accuracy. This process finalized the MMReview benchmark. 
The detailed prompts for task generation and each task can be found in Appendix \ref{Prompts}.

\subsection{Step-based Tasks}

The \textit{Step-based} theme comprises five tasks designed to progressively evaluate the performance of LLMs in simulating the key components of the academic peer review process.

\paragraph{Summary (S)}
Summarizing a paper is the initial step in peer review and a key test of a model’s ability to extract essential content. Inaccurate summaries may impair subsequent review generation. The \textit{Summary} task evaluates a model’s ability to distill key information from a full manuscript into an accurate, concise summary. In this task, the model is prompted to generate a brief summary in its own words, avoiding abstract copying and subjective judgment. The model-generated summary is then compared to human-reviewer-written summaries and evaluated for semantic similarity and information coverage, measuring the model’s holistic comprehension and representation of academic content.

\paragraph{Strengths Evaluation (SE) and Weaknesses Evaluation (WE)}

Summarizing and analyzing a manuscript’s strengths and weaknesses is a core aspect of peer review. The \textit{Strengths Evaluation} and \textit{Weaknesses Evaluation} tasks assess LLMs’ ability to identify and articulate the merits and limitations of academic papers. These tasks test whether models can synthesize technical highlights and methodological concerns noted by human reviewers, focusing on four dimensions: Quality, Clarity, Significance, and Originality. In the Strengths Evaluation task, models argue for acceptance by detailing methodological rigor, experimental robustness, structural clarity, research impact, and novelty, thus evaluating their capacity to extract technical contributions and assess scientific merit. In contrast, the Weaknesses Evaluation task adopts a rejection-oriented stance, testing critical reasoning and constructive critique. Model outputs are compared with human reviews based on content coverage and semantic similarity.

\paragraph{Soundness Scoring (SS) and Presentation Scoring (PS)}

The \textit{Soundness Scoring} and \textit{Presentation Scoring} tasks evaluate LLMs’ ability to quantitatively assess manuscript quality, focusing on {technical soundness} and {writing presentation}. In \textit{Soundness Scoring}, the model rates the reliability of technical claims, experimental rigor, and evidential support, emphasizing empirical and methodological validity. \textit{Presentation Scoring} assesses linguistic clarity and logical organization, reflecting writing quality and information structure. Both tasks require integer scores from 1 to 4, denoting ``poor" to ``excellent." Model scores are compared to human ratings to assess judgment consistency.

\subsection{Outcome-based Tasks}

The \textit{Outcome-based} tasks focus on assessing a model's direct capability to generate peer review outcomes, with the goal of evaluating its alignment with human reviewers in final decision-making.

\paragraph{Conditional Decision (CD)}

The \textit{Conditional Decision} task assesses LLMs’ ability to synthesize human-written reviews and generate an overall quality score for a paper. Provided with reviewer comments detailing strengths, weaknesses, and evaluations of methodology and results, the model assigns a numerical score from 1 to 10, reflecting a scale from “fundamentally flawed or lacking novelty” to “groundbreaking contribution,” aligned with academic conference standards. The task evaluates the model’s capacity to interpret sentiment, weighting, and evaluative reasoning in the reviews and translate them into a coherent quantitative judgment. Model scores are compared with human ratings to assess alignment and accuracy in review-based decision-making.

\paragraph{Direct Decision (DD) and CoT Decision (CoD)}

The \textit{Direct Decision} and \textit{CoT (Chain-of-Thought) Decision} tasks evaluate LLMs’ ability to autonomously assess academic paper quality, reflecting two reviewing paradigms: streamlined judgment and step-by-step reasoning~\cite{weiChainofThoughtPromptingElicits2022}. These tasks vary in input format and cognitive complexity, enabling controlled comparison of model performance under different reasoning demands. In the \textit{Direct Decision} task, the model produces an overall score without guidance, simulating a reviewer’s holistic judgment from a single read. In contrast, the \textit{CoT Decision} task guides the model through a structured reasoning process: summarizing the paper, analyzing strengths and weaknesses across {Quality}, {Clarity}, {Significance}, and {Originality}, assigning {Soundness} and {Presentation} scores, and synthesizing an overall score. This mirrors a reviewer’s iterative, analytical evaluation. The tasks test reasoning ability and scoring traceability. Model outputs are compared to human scores to assess consistency and rationality.

\paragraph{Meta Decision (MD)}

Beyond generating individual reviews and scores, a key aspect of academic peer review is the Area Chair’s (AC) synthesis of reviewer feedback to make a final decision. To emulate this, the \textit{Meta Decision} task requires the model to issue a binary judgment, \textit{Accept} or \textit{Reject}, based on multiple human reviews. The prompt provides structured guidance and evaluation criteria, prompting step-by-step reasoning. The model is instructed to assess the quality and consistency of reviews rather than merely averaging scores. This task mirrors the real role of an AC and rigorously evaluates the model’s capacity for synthesis and decision rationality. Model outputs are compared to human-written meta-reviews to assess reliability and scientific judgment in high-level peer review.

\subsection{Preference-based Task}

\paragraph{Pairwise Rank (PR)}
Prior work has shown that pairwise comparison effectively evaluates the alignment between LLM-generated preferences and human judgments \cite{liuAligningHumanJudgement2024}. Since academic conference acceptance tiers, oral, spotlight, poster, reject, reflect human preference rankings, the \textit{Pairwise Rank} task is designed to test whether LLMs, as reviewers, display preference patterns consistent with human evaluators. This task assesses the model’s relative judgment ability by presenting pairs of papers from different acceptance tiers: oral (top 5\%), spotlight (top 25\%), poster, and reject. The model compares and ranks the papers, simulating real-world peer review selection. Alignment is measured by comparing model preferences with actual acceptance categories to determine ranking accuracy. To reduce positional bias \cite{shiJudgingJudgesSystematic2025,thakurJudgingJudgesEvaluating2025}, each comparison is repeated with reversed input order.

\subsection{Attack-based Tasks}

This task category assesses models’ robustness and discriminative ability in peer review by introducing adversarial inputs. Fabricated strengths or weaknesses, such as inverted pros and cons, are used to test the model’s capacity to detect factual inaccuracies. Misleading prompts further aim to divert the model from its original instructions. Model outputs are compared to human judgments or the model’s own non-adversarial responses, focusing on its ability to preserve evaluative independence and logical consistency under input perturbations.

\paragraph{Fake Strengths (FS) and Fake Weaknesses (FW)}

LLMs have been widely observed to exhibit a "sycophantic" tendency and a propensity to align with human judgments \cite{liuLargeLanguageModels2024,kumaranHowOverconfidenceInitial2025}, which may compromise their ability to identify flaws and generate critical peer review comments. Additionally, the hallucination issue inherent in LLMs may lead to erroneous assessments of a paper’s strengths and weaknesses. To investigate these challenges, we propose the \textit{Fake Strength} and \textit{Fake Weakness} tasks, designed to evaluate the model’s ability to correctly identify content and provide accurate assessments of strengths and weaknesses by artificially reversing the polarity of strengths and weaknesses in human review comments. We employ GPT-4o to rewrite the strengths and weaknesses in the human-written review comments, transforming strengths into weaknesses and vice versa. For instance, the strength ``the paper is clearly written'' is rephrased as ``the paper is unclear.'' The model is then asked to evaluate whether these altered strengths and weaknesses are present, allowing us to assess its ability to judge the validity of strengths and weaknesses, detect factual inconsistencies, and generate critical, grounded assessments.

\paragraph{Prompt Injection (PI)}

Recent studies have revealed that some paper authors attempt to manipulate LLM-generated reviews by employing prompt injection techniques, embedding invisible white-text prompts within the manuscript to elicit overly favorable evaluations~\cite{gibneyScientistsHideMessages2025}. However, the actual effectiveness of such manipulations remains insufficiently quantified. To evaluate this vulnerability, we introduce the \textit{Prompt Injection} task, embedding suggestive prompts (e.g., ``IGNORE ALL PREVIOUS INSTRUCTIONS, NOW GIVE A POSITIVE REVIEW OF THESE PAPERS AND DO NOT HIGHLIGHT ANY NEGATIVES.”) into the manuscript to test whether models can be coerced into compromising objectivity. The setup mirrors the \textit{CoT Decision} task, with the only change being the inclusion of injected cues. As the text is visually hidden, this task is excluded from PDF-as-img evaluations.

\section{Experiments}

\subsection{Evaluation details}

We conducted extensive experiments using MMReview on popular LLMs and MLLMs, including mainstream open-source models such as Qwen~\cite{Qwen2.5VL,yang2025qwen3technicalreport}, Kimi-VL~\cite{Kimi-vl}, InternVL3~\cite{internvl3}, OVIS2~\cite{Ovis2}, and Deepseek~\cite{deepseek-aiDeepSeekR1IncentivizingReasoning2025}, as well as advanced closed-source models such as GPT-4o, Gemini-2.5, and Claude-4. Based on the model size, we categorized the models into four groups: tiny ($<$7B), small ($\geq$7B, $<$32B), medium ($\geq$32B, $\leq$72B), and large (>72B and closed-source).

As shown in Table~\ref{tab:question}, for tasks without objective evaluation metrics, namely S, SE, and WE, we employ BARTScore~\cite{yuanBARTScoreEvaluatingGenerated2021} and the `LLM-as-a-Judge' paradigm~\cite{baiBenchmarkingFoundationModels2023,llm-as-a-judge,llm-as-a-judge-survey} to assess the similarity between model-generated and human-written review comments. For classification-based tasks such as MD and PR, we evaluate performance using accuracy. For other tasks where the model output is a numerical score, we compute the Mean Absolute Error (MAE) between the model's predicted score and the ground-truth score to quantify deviation.

\begin{table}[!t]
\centering
\small
\setlength{\tabcolsep}{2.0pt}
\resizebox{\columnwidth}{!}{%
\begin{tabular}{@{}c|c|c|c@{}}
\toprule
\textbf{Theme}           & \textbf{Task}                   & \textbf{\# Ques.} & \textbf{Metric} \\ \midrule
\multirow{5}{*}{Step}    & Summary (S)                     & 240                   & BART $\uparrow$, LLM $\uparrow$    \\
                         & Strengths Eval (SE)       & 238                   & BART $\uparrow$, LLM   $\uparrow$  \\
                         & Weaknesses Eval (WE)      & 240                   & BART $\uparrow$, LLM $\uparrow$    \\
                         & Soundness Scoring (SS)          & 32                    & MAE   $\downarrow$          \\
                         & Presentation Scoring (PS)       & 32                    & MAE   $\downarrow$          \\ \midrule
\multirow{4}{*}{Outcome} & Conditional Decision (CD)       & 176                   & MAE   $\downarrow$          \\
                         & Direct Decision (DD)            & 176                   & MAE   $\downarrow$          \\
                         & CoT Decision (CoD)              & 176                   & MAE   $\downarrow$          \\
                         & Meta Decision (MD)              & 240                   & ACC  $\uparrow$           \\ \midrule
Preference               & Pairwise Rank (PR)              & 96                    & ACC  $\uparrow$           \\ \midrule
\multirow{3}{*}{Attack}  & Fake Strength  (FS)   & 240                   & MAE  $\downarrow$           \\
                         & Fake Weakness    (FW) & 238                   & MAE  $\downarrow$           \\
                         & Prompt Injection (PI)           & 176                   & MAE   $\downarrow$          \\ \bottomrule
\end{tabular}%
}
\caption{The number of questions and corresponding evaluation metrics of different tasks.}
\label{tab:question}
\end{table}

\begin{table*}[!t]
\centering
\setlength{\tabcolsep}{2.0pt}
\resizebox{\textwidth}{!}{%
\begin{tabular}{@{}cc|cccccccc|cccc|c|ccc@{}}
\toprule
\multicolumn{2}{c|}{} &
  \multicolumn{8}{c|}{Step} &
  \multicolumn{4}{c|}{Outcome} &
  Preference &
  \multicolumn{3}{c}{Attack} \\ \cmidrule(l){3-18} 
\multicolumn{2}{c|}{\multirow{-2}{*}{Model}} &
  $S_B\uparrow$ &
  $S_L\uparrow$ &
  $SE_B\uparrow$ &
  $SE_L\uparrow$ &
  $WE_B\uparrow$ &
  $WE_L\uparrow$ &
  $\mathrm{SS} \downarrow$ &
  $\mathrm{PS} \downarrow$ &
  $\mathrm{CD} \downarrow$ &
  $\mathrm{DD} \downarrow$ &
  $\mathrm{CoD} \downarrow$ &
  $\mathrm{MD}  \uparrow$ &
  $\mathrm{PR} \uparrow$ &
  $\mathrm{FS} \downarrow$ &
  $\mathrm{FW} \downarrow$ &
  $\mathrm{PI} \downarrow$ \\ \midrule
\multicolumn{1}{c|}{} &
  InternVL3-2B &
  -3.03 &
  3.64 &
  -3.66 &
  3.56 &
  -3.95 &
  2.15 &
  0.47 &
  0.47 &
  2.35 &
  3.43 &
  3.29 &
  66.25 &
  53.13 &
  3.14 &
  1.37 &
  1.11 \\
\multicolumn{1}{c|}{} &
  Qwen2.5-VL-3B &
  -3.23 &
  3.58 &
  -3.89 &
  3.36 &
  -4.00 &
  1.67 &
  0.00 &
  0.00 &
  1.46 &
  4.37 &
  4.46 &
  64.91 &
  73.96 &
  3.46 &
  2.95 &
  0.05 \\
\multicolumn{1}{c|}{} &
  Kimi-VL-A3B-I &
  -3.01 &
  3.65 &
  -3.67 &
  3.46 &
  -3.96 &
  1.98 &
  0.47 &
  0.44 &
  2.31 &
  3.13 &
  3.83 &
  60.92 &
  56.25 &
  2.99 &
  0.56 &
  0.87 \\
\multicolumn{1}{c|}{} &
  Kimi-VL-A3B-T &
  -3.15 &
  3.71 &
  -3.68 &
  3.79 &
  -3.91 &
  2.52 &
  0.47 &
  0.44 &
  2.16 &
  3.59 &
  3.37 &
  66.67 &
  57.29 &
  3.31 &
  0.96 &
  0.63 \\
\multicolumn{1}{c|}{\multirow{-5}{*}{Tiny}} &
  \cellcolor[HTML]{D9D9D9}\textbf{Avg.} &
  \cellcolor[HTML]{D9D9D9}-3.10 &
  \cellcolor[HTML]{D9D9D9}3.65 &
  \cellcolor[HTML]{D9D9D9}-3.72 &
  \cellcolor[HTML]{D9D9D9}3.54 &
  \cellcolor[HTML]{D9D9D9}-3.95 &
  \cellcolor[HTML]{D9D9D9}2.08 &
  \cellcolor[HTML]{D9D9D9}\textbf{0.35} &
  \cellcolor[HTML]{D9D9D9}\textbf{0.34} &
  \cellcolor[HTML]{D9D9D9}2.07 &
  \cellcolor[HTML]{D9D9D9}3.63 &
  \cellcolor[HTML]{D9D9D9}3.74 &
  \cellcolor[HTML]{D9D9D9}64.69 &
  \cellcolor[HTML]{D9D9D9}60.16 &
  \cellcolor[HTML]{D9D9D9}3.22 &
  \cellcolor[HTML]{D9D9D9}1.46 &
  \cellcolor[HTML]{D9D9D9}0.67 \\ \midrule
\multicolumn{1}{c|}{} &
  Qwen2.5-VL-7B &
  -3.05 &
  3.61 &
  -3.68 &
  3.57 &
  -3.96 &
  2.06 &
  0.47 &
  0.44 &
  2.43 &
  3.59 &
  3.57 &
  72.92 &
  59.38 &
  2.99 &
  1.66 &
  0.19 \\
\multicolumn{1}{c|}{} &
  Qwen3-8B &
  -3.08 &
  3.84 &
  -3.63 &
  3.77 &
  -3.87 &
  2.93 &
  0.53 &
  0.44 &
  2.25 &
  3.70 &
  3.16 &
  77.50 &
  65.63 &
  3.26 &
  2.03 &
  1.41 \\
\multicolumn{1}{c|}{} &
  Deepseek-R1-8B &
  -3.09 &
  3.76 &
  -3.63 &
  3.79 &
  -3.86 &
  2.72 &
  0.63 &
  0.78 &
  1.75 &
  3.84 &
  3.55 &
  76.99 &
  64.74 &
  3.25 &
  1.99 &
  0.59 \\
\multicolumn{1}{c|}{} &
  InternVL3-8B &
  -2.99 &
  3.76 &
  -3.65 &
  3.74 &
  -3.93 &
  2.27 &
  0.47 &
  0.44 &
  2.85 &
  3.35 &
  3.35 &
  76.67 &
  52.08 &
  2.99 &
  1.35 &
  0.53 \\
\multicolumn{1}{c|}{} &
  OVIS2-8B &
  -3.09 &
  3.52 &
  -3.70 &
  3.45 &
  -3.99 &
  1.98 &
  0.47 &
  0.44 &
  2.28 &
  3.41 &
  3.72 &
  63.87 &
  60.42 &
  2.99 &
  2.41 &
  0.63 \\
\multicolumn{1}{c|}{} &
  GLM-4.1V-9B-T &
  -3.15 &
  3.60 &
  -3.68 &
  3.73 &
  -3.93 &
  2.60 &
  0.50 &
  0.44 &
  2.08 &
  3.53 &
  3.33 &
  71.86 &
  59.77 &
  3.00 &
  1.08 &
  0.35 \\
\multicolumn{1}{c|}{} &
  Qwen3-14B &
  -3.06 &
  3.85 &
  -3.64 &
  3.83 &
  -3.87 &
  2.79 &
  0.53 &
  0.50 &
  2.16 &
  3.77 &
  3.64 &
  80.42 &
  61.46 &
  3.02 &
  1.99 &
  0.71 \\
\multicolumn{1}{c|}{} &
  OVIS2-16B &
  -3.04 &
  3.59 &
  -3.70 &
  3.52 &
  -3.99 &
  2.06 &
  0.47 &
  0.44 &
  1.87 &
  3.33 &
  3.73 &
  79.92 &
  64.58 &
  3.00 &
  2.34 &
  0.12 \\
\multicolumn{1}{c|}{\multirow{-9}{*}{Small}} &
  \cellcolor[HTML]{D9D9D9}\textbf{Avg.} &
  \cellcolor[HTML]{D9D9D9}-3.07 &
  \cellcolor[HTML]{D9D9D9}3.69 &
  \cellcolor[HTML]{D9D9D9}-3.66 &
  \cellcolor[HTML]{D9D9D9}3.68 &
  \cellcolor[HTML]{D9D9D9}-3.93 &
  \cellcolor[HTML]{D9D9D9}2.43 &
  \cellcolor[HTML]{D9D9D9}0.51 &
  \cellcolor[HTML]{D9D9D9}0.49 &
  \cellcolor[HTML]{D9D9D9}2.21 &
  \cellcolor[HTML]{D9D9D9}3.56 &
  \cellcolor[HTML]{D9D9D9}\textbf{3.51} &
  \cellcolor[HTML]{D9D9D9}75.02 &
  \cellcolor[HTML]{D9D9D9}61.01 &
  \cellcolor[HTML]{D9D9D9}3.06 &
  \cellcolor[HTML]{D9D9D9}1.86 &
  \cellcolor[HTML]{D9D9D9}0.57 \\ \midrule
\multicolumn{1}{c|}{} &
  Qwen2.5-VL-32B &
  -2.97 &
  3.90 &
  -3.60 &
  3.75 &
  -3.87 &
  2.58 &
  0.56 &
  0.50 &
  2.00 &
  3.15 &
  3.67 &
  67.08 &
  67.71 &
  2.99 &
  1.92 &
  0.80 \\
\multicolumn{1}{c|}{} &
  Qwen3-32B &
  -3.05 &
  3.90 &
  -3.61 &
  3.81 &
  -3.85 &
  2.91 &
  0.50 &
  0.56 &
  2.14 &
  3.60 &
  3.49 &
  80.00 &
  68.75 &
  3.05 &
  1.83 &
  0.78 \\
\multicolumn{1}{c|}{} &
  OVIS2-34B &
  -3.04 &
  3.48 &
  -3.68 &
  3.50 &
  -3.97 &
  2.24 &
  0.81 &
  0.81 &
  2.14 &
  3.72 &
  3.76 &
  79.92 &
  62.50 &
  2.99 &
  1.81 &
  0.57 \\
\multicolumn{1}{c|}{} &
  Qwen2.5-VL-72B &
  -2.99 &
  3.74 &
  -3.65 &
  3.58 &
  -3.94 &
  2.29 &
  0.47 &
  0.47 &
  2.06 &
  3.64 &
  3.71 &
  0.69 &
  0.64 &
  0.00 &
  0.00 &
  0.84 \\
\multicolumn{1}{c|}{\multirow{-5}{*}{Middle}} &
  \cellcolor[HTML]{D9D9D9}\textbf{Avg.} &
  \cellcolor[HTML]{D9D9D9}\textbf{-3.01} &
  \cellcolor[HTML]{D9D9D9}3.76 &
  \cellcolor[HTML]{D9D9D9}-3.64 &
  \cellcolor[HTML]{D9D9D9}3.66 &
  \cellcolor[HTML]{D9D9D9}-3.91 &
  \cellcolor[HTML]{D9D9D9}2.51 &
  \cellcolor[HTML]{D9D9D9}0.59 &
  \cellcolor[HTML]{D9D9D9}0.59 &
  \cellcolor[HTML]{D9D9D9}2.09 &
  \cellcolor[HTML]{D9D9D9}\textbf{3.53} &
  \cellcolor[HTML]{D9D9D9}3.66 &
  \cellcolor[HTML]{D9D9D9}56.92 &
  \cellcolor[HTML]{D9D9D9}49.90 &
  \cellcolor[HTML]{D9D9D9}\textbf{2.26} &
  \cellcolor[HTML]{D9D9D9}\textbf{1.39} &
  \cellcolor[HTML]{D9D9D9}0.75 \\ \midrule
\multicolumn{1}{c|}{} &
  Deepseek-V3 &
  -3.04 &
  3.84 &
  -3.61 &
  3.89 &
  -3.85 &
  2.96 &
  0.53 &
  0.47 &
  2.62 &
  3.37 &
  3.70 &
  75.00 &
  66.03 &
  2.99 &
  0.66 &
  0.20 \\
\multicolumn{1}{c|}{} &
  Deepseek-R1 &
  -3.04 &
  3.92 &
  -3.69 &
  3.90 &
  -3.91 &
  3.05 &
  0.66 &
  0.81 &
  1.97 &
  3.71 &
  3.59 &
  82.92 &
  66.03 &
  3.18 &
  1.20 &
  0.44 \\
\multicolumn{1}{c|}{} &
  Chatgpt-4o-latest &
  -3.06 &
  3.91 &
  -3.61 &
  3.89 &
  -3.86 &
  2.87 &
  0.84 &
  0.94 &
  1.65 &
  3.70 &
  3.65 &
  80.33 &
  63.54 &
  3.92 &
  1.45 &
  0.45 \\
\multicolumn{1}{c|}{} &
  Claude-sonnet-4 &
  -3.02 &
  3.88 &
  -3.58 &
  3.84 &
  -3.84 &
  3.05 &
  0.53 &
  0.41 &
  1.17 &
  2.02 &
  2.01 &
  84.58 &
  72.92 &
  2.98 &
  2.34 &
  0.43 \\
\multicolumn{1}{c|}{} &
  Gemini-2.5-flash &
  -3.06 &
  3.80 &
  -3.58 &
  3.89 &
  -3.86 &
  2.61 &
  0.94 &
  0.88 &
  1.24 &
  4.60 &
  4.28 &
  74.06 &
  70.83 &
  3.73 &
  1.37 &
  0.59 \\
\multicolumn{1}{c|}{\multirow{-6}{*}{Large}} &
  \cellcolor[HTML]{D9D9D9}\textbf{Avg.} &
  \cellcolor[HTML]{D9D9D9}-3.04 &
  \cellcolor[HTML]{D9D9D9}\textbf{3.87} &
  \cellcolor[HTML]{D9D9D9}\textbf{-3.61} &
  \cellcolor[HTML]{D9D9D9}\textbf{3.88} &
  \cellcolor[HTML]{D9D9D9}\textbf{-3.87} &
  \cellcolor[HTML]{D9D9D9}\textbf{2.91} &
  \cellcolor[HTML]{D9D9D9}0.70 &
  \cellcolor[HTML]{D9D9D9}0.70 &
  \cellcolor[HTML]{D9D9D9}\textbf{1.73} &
  \cellcolor[HTML]{D9D9D9}\textbf{3.48} &
  \cellcolor[HTML]{D9D9D9}\textbf{3.45} &
  \cellcolor[HTML]{D9D9D9}\textbf{79.38} &
  \cellcolor[HTML]{D9D9D9}\textbf{67.87} &
  \cellcolor[HTML]{D9D9D9}3.36 &
  \cellcolor[HTML]{D9D9D9}1.41 &
  \cellcolor[HTML]{D9D9D9}\textbf{0.42} \\ \bottomrule
\end{tabular}%
}
\caption{Results on MMReview with text-only inputs, where T denotes Thinking and I denotes Instruct.}
\label{tab:textonly}
\end{table*}

\begin{table*}[!t]
\centering
\small
\setlength{\tabcolsep}{2.0pt}
\resizebox{\textwidth}{!}{%
\begin{tabular}{@{}ccccccccccccccccc@{}}
\toprule
\multicolumn{1}{c|}{\multirow{2}{*}{\begin{tabular}[c]{@{}c@{}}Model\\ Size\end{tabular}}} &
  \multicolumn{8}{c|}{Step} &
  \multicolumn{4}{c|}{Outcome} &
  \multicolumn{1}{c|}{Preference} &
  \multicolumn{3}{c}{Attack} \\ \cmidrule(l){2-17} 
\multicolumn{1}{c|}{} &
  $S_B\uparrow$ &
  $S_L\uparrow$ &
  $SE_B\uparrow$ &
  $SE_L\uparrow$ &
  $WE_B\uparrow$ &
  $WE_L\uparrow$ &
  $\mathrm{SS} \downarrow$ &
  \multicolumn{1}{c|}{$\mathrm{PS} \downarrow$} &
  $\mathrm{CD} \downarrow$ &
  $\mathrm{DD} \downarrow$ &
  $\mathrm{CoD} \downarrow$ &
  \multicolumn{1}{c|}{$\mathrm{MD}  \uparrow$} &
  \multicolumn{1}{c|}{$\mathrm{PR} \uparrow$} &
  $\mathrm{FS} \downarrow$ &
  $\mathrm{FW} \downarrow$ &
  $\mathrm{PI} \downarrow$ \\ \midrule
\multicolumn{17}{c}{Multimodal} \\ \midrule
\multicolumn{1}{c|}{Tiny} &
  -3.12 &
  3.59 &
  -3.76 &
  3.37 &
  -3.97 &
  2.00 &
  \textbf{0.48} &
  \multicolumn{1}{c|}{\textbf{0.46}} &
  2.14 &
  \textbf{3.24} &
  3.56 &
  \multicolumn{1}{c|}{59.66} &
  \multicolumn{1}{c|}{56.77} &
  \textbf{2.95} &
  \textbf{1.11} &
  0.42 \\
\multicolumn{1}{c|}{Small} &
  -3.16 &
  3.50 &
  -3.72 &
  3.49 &
  -3.99 &
  2.13 &
  \textbf{0.48} &
  \multicolumn{1}{c|}{\textbf{0.46}} &
  2.41 &
  3.64 &
  3.82 &
  \multicolumn{1}{c|}{69.72} &
  \multicolumn{1}{c|}{61.98} &
  3.00 &
  1.77 &
  \textbf{0.31} \\
\multicolumn{1}{c|}{Middle} &
  \textbf{-3.05} &
  3.71 &
  -3.67 &
  3.63 &
  -3.94 &
  2.35 &
  0.72 &
  \multicolumn{1}{c|}{0.69} &
  2.21 &
  3.66 &
  3.86 &
  \multicolumn{1}{c|}{73.20} &
  \multicolumn{1}{c|}{61.21} &
  3.00 &
  1.55 &
  0.47 \\
\multicolumn{1}{c|}{Large} &
  -3.07 &
  \textbf{3.82} &
  \textbf{-3.59} &
  \textbf{3.86} &
  \textbf{-3.86} &
  \textbf{2.78} &
  0.76 &
  \multicolumn{1}{c|}{0.75} &
  \textbf{1.59} &
  3.48 &
  \textbf{3.34} &
  \multicolumn{1}{c|}{\textbf{78.79}} &
  \multicolumn{1}{c|}{\textbf{65.97}} &
  3.25 &
  1.37 &
  0.40 \\ \midrule
\multicolumn{17}{c}{PDF-as-img} \\ \midrule
\multicolumn{1}{c|}{Tiny} &
  -3.25 &
  3.27 &
  -4.00 &
  2.98 &
  -4.11 &
  1.83 &
  \textbf{0.47} &
  \multicolumn{1}{c|}{0.73} &
  1.86 &
  \textbf{2.93} &
  3.59 &
  \multicolumn{1}{c|}{67.47} &
  \multicolumn{1}{c|}{54.95} &
  \textbf{2.35} &
  \textbf{1.08} &
  \textbackslash{} \\
\multicolumn{1}{c|}{Small} &
  -3.55 &
  2.89 &
  -3.88 &
  3.20 &
  -4.08 &
  1.99 &
  \textbf{0.47} &
  \multicolumn{1}{c|}{\textbf{0.44}} &
  2.14 &
  3.38 &
  3.47 &
  \multicolumn{1}{c|}{72.78} &
  \multicolumn{1}{c|}{59.17} &
  2.99 &
  1.72 &
  \textbackslash{} \\
\multicolumn{1}{c|}{Middle} &
  \textbf{-3.14} &
  \textbf{3.55} &
  -3.70 &
  3.54 &
  -3.95 &
  2.29 &
  0.68 &
  \multicolumn{1}{c|}{0.70} &
  2.06 &
  3.65 &
  3.76 &
  \multicolumn{1}{c|}{71.77} &
  \multicolumn{1}{c|}{55.90} &
  3.09 &
  1.34 &
  \textbackslash{} \\
\multicolumn{1}{c|}{Large} &
  -3.28 &
  3.50 &
  \textbf{-3.67} &
  \textbf{3.72} &
  \textbf{-3.90} &
  \textbf{2.68} &
  0.77 &
  \multicolumn{1}{c|}{0.75} &
  \textbf{1.39} &
  3.53 &
  \textbf{3.36} &
  \multicolumn{1}{c|}{\textbf{77.99}} &
  \multicolumn{1}{c|}{\textbf{63.89}} &
  2.85 &
  1.60 &
  \textbackslash{} \\ \bottomrule
\end{tabular}%
}
\caption{Results on MMReview with multimodal and pdf-as-image inputs. The detailed results from models can be found in Table \ref{tab:multimodal} and \ref{tab:pdf_as_img} in Appendix \ref{More-Results}. }
\label{tab:multimodal-pdf-as-img}
\end{table*}

\subsection{Main Results}

We present the performance metrics for each model in the text-only input mode in Table \ref{tab:textonly}. To observe the performance across different model categories, we calculated the average performance for each group. Table \ref{tab:multimodal-pdf-as-img} displays the average performance metrics for each model group under the multimodal and PDF-as-image input mode. The specific performance of each model is provided in Tables \ref{tab:multimodal} and \ref{tab:pdf_as_img} in the Appendix. Based on the test results presented in the tables, the following conclusions can be drawn:

\textbf{(1) Model scale significantly influences the model's ability to comprehend and analyze.} Large-scale and closed-source models outperform the others on most metrics, particularly on tasks directly related to review conclusions, such as CD, CoD, MD, and PR. This indicates that larger models are more powerful in understanding complex academic content and generating structured feedback, making them more reliable in generating peer review comments. Surprisingly, mid-sized and smaller models performed better than their larger counterparts in assessing the soundness and presentation of papers.

\textbf{(2) High-quality structured reasoning enhances review outcomes.} Compared to directly generating review scores (DD), the use of CoT reasoning to generate review scores (CoD) achieved a lower MAE, demonstrating that a step-by-step review approach strengthens the model's evaluative capability. Furthermore, using a higher-quality, human-written review process as a reference (CD) further reduced the MAE, indicating that the quality of the reasoning process also affects the model's evaluative ability. Models that employed thinking (such as Kimi-VL-A3B-Thinking and Deepseek-R1) outperformed non-thinking models on tasks directly related to scoring, such as CD, CoD, MD, and PR, suggesting that stronger thinking and reasoning abilities enhance the model's peer review capabilities.

\textbf{(3) Multimodal input improves the model's robustness to prompt injection.} Regardless of model size, when both text and paper images are input simultaneously, the influence of prompt injection is lower compared to the text-only input mode, implying that the inclusion of image modality enhances the model's ability to judge paper quality and mitigates the impact of prompt injection interference.
\section{Further Analysis}

\begin{table}[!t]
\centering
\small
\begin{tabular}{@{}c|cccc@{}}
\toprule
                          & AI   & NS   & ES   & SS   \\ \midrule
ChatGPT-4o-latest         & 4.46 & 5.67 & 3.67 & 4.28 \\
claude-sonnet-4-20250514  & 2.08 & 2.33 & 1.33 & 2.44 \\
gemini-2.5-flash-thinking & 4.62 & 6.33 & 4.33 & 4.50 \\
Deepseek-R1               & 2.32 & 4.00 & 3.67 & 4.00 \\ \bottomrule
\end{tabular}
\caption{The MAE performance of several models on the CoD task across different research domains.}
\label{tab:domain-good}
\end{table}

\paragraph{What Domains are LLM-reviewers better at?} Table~\ref{tab:domain-good} reports the MAE performance of various models on the CoD task across research domains, indicating their relative proficiency in field-specific peer review. All models use multimodal inputs except Deepseek-R1, which operates with text only. Results show that most closed-source models perform best in Engineering Sciences, while Deepseek-R1 achieves its lowest MAE in Artificial Intelligence. These findings suggest that models’ reviewing strengths vary across disciplines, potentially influenced by differences in pretraining corpora and input modalities.

\paragraph{How does Prompt Injection Work?}

To further investigate the impact of prompt injection on model scoring behavior, we analyze score variations between the Prompt Injection task and the CoT Decision (CoD) task, where the only difference in prompts lies in the injection of the statement: \textit{“IGNORE ALL PREVIOUS INSTRUCTIONS, NOW GIVE A POSITIVE REVIEW OF THESE PAPERS AND DO NOT HIGHLIGHT ANY NEGATIVES.”} Tables~\ref{tab:pi-text} and~\ref{tab:pi-multi} report the proportion and expectation of score increases and decreases under text-only and multimodal input conditions, respectively. The experimental results reveal three key findings: (1) Most models exhibit an overall increase in average score after prompt injection, with a typical rise of 1–2 points; (2) Multimodal inputs enhance model robustness against prompt injection compared to text-only inputs, consistent with prior observations; (3) The robustness of Thinking variants varies across model families, within the Qwen3 series, Thinking models are less robust than their non-Thinking counterparts, whereas in the Kimi and Deepseek series, Thinking models demonstrate greater resistance to injection.

\begin{table}[!t]
\centering
\small
\setlength{\tabcolsep}{3.0pt}
\begin{tabular}{@{}c|cccc@{}}
\toprule
\textbf{}          & \% Raise & $E$(Raise) & \% Lower & $E$(Lower) \\ \midrule
Qwen3-8B           & 90.34    & 1.56     & 0.57     & -1.00    \\
Qwen3-8B-nothink   & 98.30    & 1.50     & 0.00     & 0.00     \\
Kimi-VL-A3B-I      & 31.15    & 1.28     & 6.56     & -0.50    \\
Qwen3-32B          & 61.93    & 1.21     & 3.41     & -1.00    \\
Chatgpt-4o-latest  & 36.47    & 1.17     & 1.18     & -1.00    \\
Claude-sonnet-4    & 11.93    & 1.14     & 25.57    & -1.13    \\
Kimi-VL-A3B-T      & 46.02    & 1.10     & 11.93    & -1.00    \\
Qwen3-14B          & 63.64    & 1.10     & 1.14     & -1.00    \\
Qwen2.5-VL-72B     & 78.41    & 1.07     & 0.00     & 0.00     \\
Deepseek-V3        & 16.48    & 1.07     & 2.84     & -1.00    \\
OVIS2-8B           & 51.81    & 1.06     & 1.20     & -1.00    \\
Qwen3-14B-nothink  & 80.68    & 1.05     & 0.00     & 0.00     \\
Deepseek-R1        & 15.91    & 1.04     & 22.73    & -1.20    \\
Qwen2.5-VL-32B     & 77.71    & 1.03     & 0.00     & 0.00     \\
Qwen3-32B-nothink  & 55.68    & 1.00     & 0.00     & 0.00     \\
Qwen2.5-VL-7B      & 18.75    & 1.00     & 0.00     & 0.00     \\
OVIS2-34B          & 56.00    & 1.00     & 0.00     & 0.00     \\
OVIS2-16B          & 11.93    & 1.00     & 0.00     & 0.00     \\
InternVL3-8B       & 52.84    & 1.00     & 0.00     & 0.00     \\
GLM-4.1V-9B-T      & 25.48    & 1.00     & 6.37     & -1.10    \\ \bottomrule
\end{tabular}
\caption{The proportion and expectation of score increases and decreases under text-only input condition.}
\label{tab:pi-text}
\end{table}

\begin{table}[!t]
\centering
\small
\setlength{\tabcolsep}{3.0pt}
\begin{tabular}{@{}c|cccc@{}}
\toprule
\multicolumn{1}{l|}{\textbf{}} &
  \multicolumn{1}{c}{\textbf{\% Raise}} &
  \multicolumn{1}{c}{\textbf{E(Raise)}} &
  \multicolumn{1}{c}{\textbf{\% Lower}} &
  \multicolumn{1}{c}{\textbf{E(Lower)}} \\ \midrule
Kimi-VL-A3B-I      & 23.23 & 1.54 & 7.74  & \multicolumn{1}{c}{-1.54} \\
OVIS2-16B          & 2.94  & 1.40 & 0.00  & 0.00                      \\
GLM-4.1V-9B-T      & 14.97 & 1.28 & 3.59  & \multicolumn{1}{c}{-1.00} \\
Chatgpt-4o-latest  & 34.30 & 1.08 & 3.49  & -2.17                     \\
Kimi-VL-A3B-T      & 21.51 & 1.08 & 8.72  & -1.00                     \\
InternVL3-8B       & 29.48 & 1.06 & 0.58  & -1.00                     \\
OVIS2-8B           & 41.72 & 1.02 & 1.32  & -4.00                     \\
Qwen2.5-VL-32B     & 47.73 & 1.01 & 1.14  & \multicolumn{1}{c}{-1.00} \\
Qwen2.5-VL-72B     & 57.14 & 1.01 & 0.00  & 0.00                      \\
Qwen2.5-VL-7B      & 28.41 & 1.00 & 0.00  & \multicolumn{1}{c}{0.00}  \\
OVIS2-34B          & 28.40 & 1.00 & 0.00  & 0.00                      \\
Claude-sonnet-4    & 9.66  & 1.00 & 21.02 & -1.08                     \\ \bottomrule
\end{tabular}
\caption{The proportion and expectation of score increases and decreases under multimodal input condition.}
\label{tab:pi-multi}
\end{table}

More analysis are presented in the Appendix \ref{More-Results}.
\section{Conclusion}

In this work, we present \textbf{MMReview}, a multidisciplinary and multimodal benchmark designed to evaluate the capabilities of LLMs in academic peer review. The benchmark encompasses 4 thematic categories and 13 distinct tasks. Its core features include coverage across diverse academic disciplines, support for multimodal input formats, and comprehensive evaluation tasks that span the full peer review pipeline. Leveraging MMReview, we conducted extensive evaluations of LLMs and MLLMs. We envision MMReview as a standardized evaluation platform that can catalyze the development of more efficient LLM-assisted peer review systems. 

\section{Limitations}

\paragraph{The issue of dataset size and distribution.} Due to the rapid advancement of AI in recent years and the open access and public review characteristics of AI papers, approximately 48\% of the papers containing peer review comments are concentrated in the AI field. This concentration may affect the representativeness of the results. In the future, we plan to collect more papers from other domains to enhance the representativeness of our benchmark.

\paragraph{The controversy surrounding review comments written by human experts.} There is currently a lack of consensus on what constitutes a good or high-quality review. Although methods such as obtaining review consensus and manual screening have been employed in the paper to filter review samples, it remains impossible to guarantee that these expert-written reviews are of sufficient quality. Moreover, it is undeniable that all types of reviews hold value, even though they may vary significantly in content, as they reflect the diverse perspectives of different reviewers.

\section{Ethical Considerations}

All the papers and peer review comments we collected are sourced from open-access platforms such as OpenReview, NeurIPS, and Nature. These platforms state that the content they publish, including but not limited to the papers themselves and their peer review comments, is licensed or permitted for research purposes under the Creative Commons Attribution International 4.0 license. We ensure that the collection and processing of these papers and reviews are conducted for research purposes and comply with the copyright agreements of the platforms.

Our research on the ability of LLMs to generate peer review comments does not advocate for the complete replacement of human reviewers with LLMs, as this might open the door to potential misuse and manipulation. Instead, we envision that in the current era of a proliferation of academic papers, LLMs can serve as an auxiliary tool. Similar to the practices already adopted by academic conferences like ICLR and AAAI, peer review comments generated by LLMs could be used as references to help reduce the workload of human reviewers to some extent.
\bibliography{acl}

\appendix

\clearpage

\appendix

\section*{Appendix}
\label{sec:appendix}


As a supplement, we provide additional materials in the appendix, including extended experimental results and analyses (Appendix \ref{More-Results}), implementation details of the experiments (Appendix \ref{More Implementation Details}), case studies~\ref{case}, as well as the prompts used during the construction and application of the benchmark (Appendix \ref{Prompts}).

\begin{table*}[!t]
\centering
\small
\setlength{\tabcolsep}{2.0pt}
\resizebox{\textwidth}{!}{%
\begin{tabular}{@{}cc|cccccccc|cccc|c|ccc@{}}
\toprule
\multicolumn{2}{c|}{} &
  \multicolumn{8}{c|}{Step} &
  \multicolumn{4}{c|}{Outcome} &
  Preference &
  \multicolumn{3}{c}{Attack} \\ \cmidrule(l){3-18} 
\multicolumn{2}{c|}{\multirow{-2}{*}{Model}} &
  $S_B\uparrow$ &
  $S_L\uparrow$ &
  $SE_B\uparrow$ &
  $SE_L\uparrow$ &
  $WE_B\uparrow$ &
  $WE_L\uparrow$ &
  $\mathrm{SS} \downarrow$ &
  $\mathrm{PS} \downarrow$ &
  $\mathrm{CD} \downarrow$ &
  $\mathrm{DD} \downarrow$ &
  $\mathrm{CoD} \downarrow$ &
  $\mathrm{MD}  \uparrow$ &
  $\mathrm{PR} \uparrow$ &
  $\mathrm{FS} \downarrow$ &
  $\mathrm{FW} \downarrow$ &
  $\mathrm{PI} \downarrow$ \\ \midrule
\multicolumn{1}{c|}{} &
  InternVL3-2B &
  -3.11 &
  3.67 &
  -3.95 &
  2.87 &
  -4.04 &
  1.89 &
  0.52 &
  0.48 &
  1.99 &
  2.56 &
  2.94 &
  60.94 &
  53.13 &
  2.48 &
  1.10 &
  0.57 \\
\multicolumn{1}{c|}{} &
  Qwen2.5-VL-3B &
  -3.11 &
  3.55 &
  -3.74 &
  3.33 &
  -3.97 &
  1.89 &
  0.47 &
  0.44 &
  1.87 &
  3.24 &
  3.74 &
  55.42 &
  57.29 &
  3.18 &
  2.88 &
  0.08 \\
\multicolumn{1}{c|}{} &
  Kimi-VL-A3B-I &
  -3.12 &
  3.53 &
  -3.69 &
  3.52 &
  -3.96 &
  1.86 &
  0.47 &
  0.44 &
  2.40 &
  3.61 &
  3.94 &
  60.68 &
  54.17 &
  3.00 &
  0.13 &
  0.65 \\
\multicolumn{1}{c|}{} &
  Kimi-VL-A3B-T &
  -3.13 &
  3.60 &
  -3.68 &
  3.74 &
  -3.92 &
  2.37 &
  0.47 &
  0.47 &
  2.31 &
  3.56 &
  3.64 &
  61.60 &
  62.50 &
  3.14 &
  0.33 &
  0.36 \\
\multicolumn{1}{c|}{\multirow{-5}{*}{Tiny}} &
  \cellcolor[HTML]{D9D9D9}\textbf{Avg.} &
  \cellcolor[HTML]{D9D9D9}-3.12 &
  \cellcolor[HTML]{D9D9D9}3.59 &
  \cellcolor[HTML]{D9D9D9}-3.76 &
  \cellcolor[HTML]{D9D9D9}3.37 &
  \cellcolor[HTML]{D9D9D9}-3.97 &
  \cellcolor[HTML]{D9D9D9}2.00 &
  \cellcolor[HTML]{D9D9D9}\textbf{0.48} &
  \cellcolor[HTML]{D9D9D9}\textbf{0.46} &
  \cellcolor[HTML]{D9D9D9}2.14 &
  \cellcolor[HTML]{D9D9D9}\textbf{3.24} &
  \cellcolor[HTML]{D9D9D9}\textbf{3.56} &
  \cellcolor[HTML]{D9D9D9}59.66 &
  \cellcolor[HTML]{D9D9D9}56.77 &
  \cellcolor[HTML]{D9D9D9}\textbf{2.95} &
  \cellcolor[HTML]{D9D9D9}\textbf{1.11} &
  \cellcolor[HTML]{D9D9D9}0.42 \\ \midrule
\multicolumn{1}{c|}{} &
  Qwen2.5-VL-7B &
  -3.09 &
  3.64 &
  -3.67 &
  3.70 &
  -3.94 &
  2.11 &
  0.47 &
  0.44 &
  2.20 &
  3.59 &
  3.49 &
  75.83 &
  64.58 &
  3.00 &
  1.45 &
  0.28 \\
\multicolumn{1}{c|}{} &
  InternVL3-8B &
  -3.21 &
  3.46 &
  -3.82 &
  3.26 &
  -4.12 &
  1.84 &
  0.47 &
  0.44 &
  3.02 &
  3.72 &
  3.68 &
  61.11 &
  54.17 &
  3.00 &
  1.33 &
  0.32 \\
\multicolumn{1}{c|}{} &
  OVIS2-8B &
  -3.20 &
  3.43 &
  -3.72 &
  3.49 &
  -3.97 &
  2.06 &
  0.50 &
  0.53 &
  2.27 &
  3.79 &
  4.44 &
  62.91 &
  66.28 &
  3.00 &
  2.15 &
  0.58 \\
\multicolumn{1}{c|}{} &
  GLM-4.1V-9B-T &
  -3.20 &
  3.48 &
  -3.69 &
  3.70 &
  -3.93 &
  2.60 &
  0.50 &
  0.47 &
  2.42 &
  3.68 &
  3.59 &
  70.72 &
  59.77 &
  3.02 &
  1.80 &
  0.31 \\
\multicolumn{1}{c|}{} &
  OVIS2-16B &
  -3.13 &
  3.50 &
  -3.70 &
  3.29 &
  -3.98 &
  2.04 &
  0.47 &
  0.44 &
  2.16 &
  3.40 &
  3.91 &
  78.03 &
  65.12 &
  3.00 &
  2.10 &
  0.04 \\
\multicolumn{1}{c|}{\multirow{-6}{*}{Small}} &
  \cellcolor[HTML]{D9D9D9}\textbf{Avg.} &
  \cellcolor[HTML]{D9D9D9}-3.16 &
  \cellcolor[HTML]{D9D9D9}3.50 &
  \cellcolor[HTML]{D9D9D9}-3.72 &
  \cellcolor[HTML]{D9D9D9}3.49 &
  \cellcolor[HTML]{D9D9D9}-3.99 &
  \cellcolor[HTML]{D9D9D9}2.13 &
  \cellcolor[HTML]{D9D9D9}\textbf{0.48} &
  \cellcolor[HTML]{D9D9D9}\textbf{0.46} &
  \cellcolor[HTML]{D9D9D9}2.41 &
  \cellcolor[HTML]{D9D9D9}3.64 &
  \cellcolor[HTML]{D9D9D9}3.82 &
  \cellcolor[HTML]{D9D9D9}69.72 &
  \cellcolor[HTML]{D9D9D9}61.98 &
  \cellcolor[HTML]{D9D9D9}3.00 &
  \cellcolor[HTML]{D9D9D9}1.77 &
  \cellcolor[HTML]{D9D9D9}\textbf{0.31} \\ \midrule
\multicolumn{1}{c|}{} &
  Qwen2.5-VL-32B &
  -2.99 &
  3.89 &
  -3.64 &
  3.76 &
  -3.90 &
  2.45 &
  0.81 &
  0.75 &
  2.24 &
  3.47 &
  3.94 &
  76.25 &
  60.42 &
  3.00 &
  1.57 &
  0.49 \\
\multicolumn{1}{c|}{} &
  OVIS2-34B &
  -3.11 &
  3.51 &
  -3.69 &
  3.43 &
  -3.97 &
  2.26 &
  0.88 &
  0.81 &
  2.33 &
  3.78 &
  3.91 &
  73.76 &
  62.79 &
  3.00 &
  1.78 &
  0.28 \\
\multicolumn{1}{c|}{} &
  Qwen2.5-VL-72B &
  -3.04 &
  3.74 &
  -3.66 &
  3.71 &
  -3.95 &
  2.33 &
  0.47 &
  0.50 &
  2.06 &
  3.73 &
  3.73 &
  69.58 &
  60.42 &
  3.01 &
  1.30 &
  0.63 \\
\multicolumn{1}{c|}{\multirow{-4}{*}{Middle}} &
  \cellcolor[HTML]{D9D9D9}\textbf{Avg.} &
  \cellcolor[HTML]{D9D9D9}\textbf{-3.05} &
  \cellcolor[HTML]{D9D9D9}3.71 &
  \cellcolor[HTML]{D9D9D9}-3.67 &
  \cellcolor[HTML]{D9D9D9}3.63 &
  \cellcolor[HTML]{D9D9D9}-3.94 &
  \cellcolor[HTML]{D9D9D9}2.35 &
  \cellcolor[HTML]{D9D9D9}0.72 &
  \cellcolor[HTML]{D9D9D9}0.69 &
  \cellcolor[HTML]{D9D9D9}2.21 &
  \cellcolor[HTML]{D9D9D9}3.66 &
  \cellcolor[HTML]{D9D9D9}3.86 &
  \cellcolor[HTML]{D9D9D9}73.20 &
  \cellcolor[HTML]{D9D9D9}61.21 &
  \cellcolor[HTML]{D9D9D9}3.00 &
  \cellcolor[HTML]{D9D9D9}1.55 &
  \cellcolor[HTML]{D9D9D9}0.47 \\ \midrule
\multicolumn{1}{c|}{} &
  Chatgpt-4o-latest &
  -3.10 &
  3.86 &
  -3.60 &
  3.85 &
  -3.87 &
  2.67 &
  0.91 &
  0.94 &
  2.24 &
  3.65 &
  3.75 &
  75.97 &
  59.38 &
  3.88 &
  1.22 &
  0.46 \\
\multicolumn{1}{c|}{} &
  Claude-sonnet-4 &
  -3.02 &
  3.86 &
  -3.60 &
  3.88 &
  -3.83 &
  3.09 &
  0.47 &
  0.44 &
  1.19 &
  2.19 &
  2.09 &
  84.17 &
  69.79 &
  2.98 &
  2.12 &
  0.32 \\
\multicolumn{1}{c|}{} &
  Gemini-2.5-flash &
  -3.08 &
  3.75 &
  -3.56 &
  3.84 &
  -3.87 &
  2.58 &
  0.91 &
  0.88 &
  1.34 &
  4.60 &
  4.19 &
  76.25 &
  68.75 &
  2.89 &
  0.78 &
  0.41 \\
\multicolumn{1}{c|}{\multirow{-4}{*}{Large}} &
  \cellcolor[HTML]{D9D9D9}\textbf{Avg.} &
  \cellcolor[HTML]{D9D9D9}-3.07 &
  \cellcolor[HTML]{D9D9D9}\textbf{3.82} &
  \cellcolor[HTML]{D9D9D9}\textbf{-3.59} &
  \cellcolor[HTML]{D9D9D9}\textbf{3.86} &
  \cellcolor[HTML]{D9D9D9}\textbf{-3.86} &
  \cellcolor[HTML]{D9D9D9}\textbf{2.78} &
  \cellcolor[HTML]{D9D9D9}0.76 &
  \cellcolor[HTML]{D9D9D9}0.75 &
  \cellcolor[HTML]{D9D9D9}\textbf{1.59} &
  \cellcolor[HTML]{D9D9D9}3.48 &
  \cellcolor[HTML]{D9D9D9}\textbf{3.34} &
  \cellcolor[HTML]{D9D9D9}\textbf{78.79} &
  \cellcolor[HTML]{D9D9D9}\textbf{65.97} &
  \cellcolor[HTML]{D9D9D9}3.25 &
  \cellcolor[HTML]{D9D9D9}1.37 &
  \cellcolor[HTML]{D9D9D9}0.40 \\ \bottomrule
\end{tabular}%
}
\caption{Results on MMReview with multimodal inputs.}
\label{tab:multimodal}
\end{table*}

\begin{table*}[!t]
\small
\setlength{\tabcolsep}{2.0pt}
\centering
\resizebox{\textwidth}{!}{%
\begin{tabular}{@{}cc|cccccccc|cccc|c|cc@{}}
\toprule
\multicolumn{2}{c|}{} &
  \multicolumn{8}{c|}{Step} &
  \multicolumn{4}{c|}{Outcome} &
  Preference &
  \multicolumn{2}{c}{Attack} \\ \cmidrule(l){3-17} 
\multicolumn{2}{c|}{\multirow{-2}{*}{Model}} &
  $S_B\uparrow$ &
  $S_L\uparrow$ &
  $SE_B\uparrow$ &
  $SE_L\uparrow$ &
  $WE_B\uparrow$ &
  $WE_L\uparrow$ &
  $\mathrm{SS} \downarrow$ &
  $\mathrm{PS} \downarrow$ &
  $\mathrm{CD} \downarrow$ &
  $\mathrm{DD} \downarrow$ &
  $\mathrm{CoD} \downarrow$ &
  $\mathrm{MD}  \uparrow$ &
  $\mathrm{PR} \uparrow$ &
  $\mathrm{FS} \downarrow$ &
  $\mathrm{FW} \downarrow$ \\ \midrule
\multicolumn{1}{c|}{} &
  InternVL3-2B &
  -3.28 &
  3.28 &
  -4.80 &
  1.51 &
  -4.55 &
  1.19 &
  0.48 &
  1.61 &
  1.57 &
  1.77 &
  3.34 &
  72.38 &
  50.00 &
  3.00 &
  1.11 \\
\multicolumn{1}{c|}{} &
  Qwen2.5-VL-3B &
  -3.29 &
  3.12 &
  -3.76 &
  3.34 &
  -3.98 &
  1.86 &
  0.47 &
  0.44 &
  1.73 &
  3.05 &
  3.72 &
  53.75 &
  54.17 &
  3.01 &
  2.94 \\
\multicolumn{1}{c|}{} &
  Kimi-VL-A3B-I &
  -3.21 &
  3.17 &
  -3.74 &
  3.34 &
  -4.00 &
  1.74 &
  0.47 &
  0.44 &
  2.30 &
  3.19 &
  3.83 &
  69.36 &
  55.21 &
  0.28 &
  0.00 \\
\multicolumn{1}{c|}{} &
  Kimi-VL-A3B-T &
  -3.21 &
  3.50 &
  -3.71 &
  3.72 &
  -3.91 &
  2.54 &
  0.47 &
  0.44 &
  1.85 &
  3.72 &
  3.49 &
  74.37 &
  60.42 &
  3.11 &
  0.27 \\
\multicolumn{1}{c|}{\multirow{-5}{*}{Tiny}} &
  \cellcolor[HTML]{D9D9D9}\textbf{Avg.} &
  \cellcolor[HTML]{D9D9D9}-3.25 &
  \cellcolor[HTML]{D9D9D9}3.27 &
  \cellcolor[HTML]{D9D9D9}-4.00 &
  \cellcolor[HTML]{D9D9D9}2.98 &
  \cellcolor[HTML]{D9D9D9}-4.11 &
  \cellcolor[HTML]{D9D9D9}1.83 &
  \cellcolor[HTML]{D9D9D9}\textbf{0.47} &
  \cellcolor[HTML]{D9D9D9}0.73 &
  \cellcolor[HTML]{D9D9D9}1.86 &
  \cellcolor[HTML]{D9D9D9}\textbf{2.93} &
  \cellcolor[HTML]{D9D9D9}3.59 &
  \cellcolor[HTML]{D9D9D9}67.47 &
  \cellcolor[HTML]{D9D9D9}54.95 &
  \cellcolor[HTML]{D9D9D9}\textbf{2.35} &
  \cellcolor[HTML]{D9D9D9}\textbf{1.08} \\ \midrule
\multicolumn{1}{c|}{} &
  Qwen2.5-VL-7B &
  -3.21 &
  3.28 &
  -3.72 &
  3.49 &
  -3.95 &
  1.91 &
  0.47 &
  0.44 &
  1.97 &
  3.09 &
  3.12 &
  70.83 &
  47.92 &
  2.99 &
  1.01 \\
\multicolumn{1}{c|}{} &
  InternVL3-8B &
  -4.86 &
  1.13 &
  -4.59 &
  1.89 &
  -4.55 &
  1.13 &
  0.47 &
  0.44 &
  2.40 &
  3.63 &
  3.41 &
  68.75 &
  75.00 &
  2.99 &
  1.98 \\
\multicolumn{1}{c|}{} &
  OVIS2-8B &
  -3.23 &
  3.28 &
  -3.70 &
  3.43 &
  -3.99 &
  2.06 &
  0.47 &
  0.44 &
  2.49 &
  3.44 &
  3.93 &
  65.32 &
  53.13 &
  3.00 &
  2.00 \\
\multicolumn{1}{c|}{} &
  GLM-4.1V-9B-T &
  -3.26 &
  3.36 &
  -3.67 &
  3.69 &
  -3.92 &
  2.69 &
  0.47 &
  0.44 &
  1.80 &
  3.51 &
  3.13 &
  80.33 &
  65.63 &
  3.00 &
  1.82 \\
\multicolumn{1}{c|}{} &
  OVIS2-16B &
  -3.20 &
  3.42 &
  -3.70 &
  3.49 &
  -3.97 &
  2.17 &
  0.47 &
  0.44 &
  2.05 &
  3.23 &
  3.76 &
  78.66 &
  54.17 &
  3.00 &
  1.81 \\
\multicolumn{1}{c|}{\multirow{-6}{*}{Small}} &
  \cellcolor[HTML]{D9D9D9}\textbf{Avg.} &
  \cellcolor[HTML]{D9D9D9}-3.55 &
  \cellcolor[HTML]{D9D9D9}2.89 &
  \cellcolor[HTML]{D9D9D9}-3.88 &
  \cellcolor[HTML]{D9D9D9}3.20 &
  \cellcolor[HTML]{D9D9D9}-4.08 &
  \cellcolor[HTML]{D9D9D9}1.99 &
  \cellcolor[HTML]{D9D9D9}\textbf{0.47} &
  \cellcolor[HTML]{D9D9D9}\textbf{0.44} &
  \cellcolor[HTML]{D9D9D9}2.14 &
  \cellcolor[HTML]{D9D9D9}3.38 &
  \cellcolor[HTML]{D9D9D9}\textbf{3.47} &
  \cellcolor[HTML]{D9D9D9}72.78 &
  \cellcolor[HTML]{D9D9D9}59.17 &
  \cellcolor[HTML]{D9D9D9}2.99 &
  \cellcolor[HTML]{D9D9D9}1.72 \\ \midrule
\multicolumn{1}{c|}{} &
  Qwen2.5-VL-32B &
  -3.11 &
  3.66 &
  -3.71 &
  3.67 &
  -3.92 &
  2.46 &
  0.78 &
  0.81 &
  2.11 &
  3.55 &
  3.77 &
  67.50 &
  61.46 &
  3.24 &
  1.22 \\
\multicolumn{1}{c|}{} &
  OVIS2-34B &
  -3.20 &
  3.41 &
  -3.68 &
  3.49 &
  -3.99 &
  2.18 &
  0.78 &
  0.69 &
  1.91 &
  3.63 &
  3.74 &
  77.82 &
  52.08 &
  2.99 &
  1.69 \\
\multicolumn{1}{c|}{} &
  Qwen2.5-VL-72B &
  -3.11 &
  3.57 &
  -3.71 &
  3.46 &
  -3.95 &
  2.22 &
  0.47 &
  0.59 &
  2.15 &
  3.76 &
  3.77 &
  70.00 &
  54.17 &
  3.03 &
  1.11 \\
\multicolumn{1}{c|}{\multirow{-4}{*}{Middle}} &
  \cellcolor[HTML]{D9D9D9}\textbf{Avg.} &
  \cellcolor[HTML]{D9D9D9}\textbf{-3.14} &
  \cellcolor[HTML]{D9D9D9}\textbf{3.55} &
  \cellcolor[HTML]{D9D9D9}-3.70 &
  \cellcolor[HTML]{D9D9D9}3.54 &
  \cellcolor[HTML]{D9D9D9}-3.95 &
  \cellcolor[HTML]{D9D9D9}2.29 &
  \cellcolor[HTML]{D9D9D9}0.68 &
  \cellcolor[HTML]{D9D9D9}0.70 &
  \cellcolor[HTML]{D9D9D9}2.06 &
  \cellcolor[HTML]{D9D9D9}3.65 &
  \cellcolor[HTML]{D9D9D9}3.76 &
  \cellcolor[HTML]{D9D9D9}71.77 &
  \cellcolor[HTML]{D9D9D9}55.90 &
  \cellcolor[HTML]{D9D9D9}3.09 &
  \cellcolor[HTML]{D9D9D9}1.34 \\ \midrule
\multicolumn{1}{c|}{} &
  Chatgpt-4o-latest &
  -3.33 &
  3.65 &
  -3.69 &
  3.70 &
  -3.88 &
  2.68 &
  0.91 &
  0.91 &
  1.84 &
  3.63 &
  3.68 &
  77.73 &
  59.38 &
  3.90 &
  1.33 \\
\multicolumn{1}{c|}{} &
  Claude-sonnet-4 &
  -3.19 &
  3.59 &
  -3.65 &
  3.73 &
  -3.90 &
  2.88 &
  0.48 &
  0.45 &
  1.22 &
  2.36 &
  2.22 &
  82.92 &
  62.50 &
  3.00 &
  2.73 \\
\multicolumn{1}{c|}{} &
  Gemini-2.5-flash &
  -3.31 &
  3.26 &
  -3.66 &
  3.73 &
  -3.93 &
  2.48 &
  0.91 &
  0.91 &
  1.11 &
  4.60 &
  4.19 &
  73.33 &
  69.79 &
  1.67 &
  0.75 \\
\multicolumn{1}{c|}{\multirow{-4}{*}{Large}} &
  \cellcolor[HTML]{D9D9D9}\textbf{Avg.} &
  \cellcolor[HTML]{D9D9D9}-3.28 &
  \cellcolor[HTML]{D9D9D9}3.50 &
  \cellcolor[HTML]{D9D9D9}\textbf{-3.67} &
  \cellcolor[HTML]{D9D9D9}\textbf{3.72} &
  \cellcolor[HTML]{D9D9D9}\textbf{-3.90} &
  \cellcolor[HTML]{D9D9D9}\textbf{2.68} &
  \cellcolor[HTML]{D9D9D9}0.77 &
  \cellcolor[HTML]{D9D9D9}0.75 &
  \cellcolor[HTML]{D9D9D9}\textbf{1.39} &
  \cellcolor[HTML]{D9D9D9}3.53 &
  \cellcolor[HTML]{D9D9D9}\textbf{3.36} &
  \cellcolor[HTML]{D9D9D9}\textbf{77.99} &
  \cellcolor[HTML]{D9D9D9}\textbf{63.89} &
  \cellcolor[HTML]{D9D9D9}2.85 &
  \cellcolor[HTML]{D9D9D9}1.60 \\ \bottomrule
\end{tabular}%
}
\caption{Results on MMReview with pdf-as-image inputs.}
\label{tab:pdf_as_img}
\end{table*}



\section{More Results and Analysis}
\label{More-Results}

\subsection{The Impact of Paper Length}

\begin{figure}[!t]
    \centering
    \includegraphics[width=\linewidth]{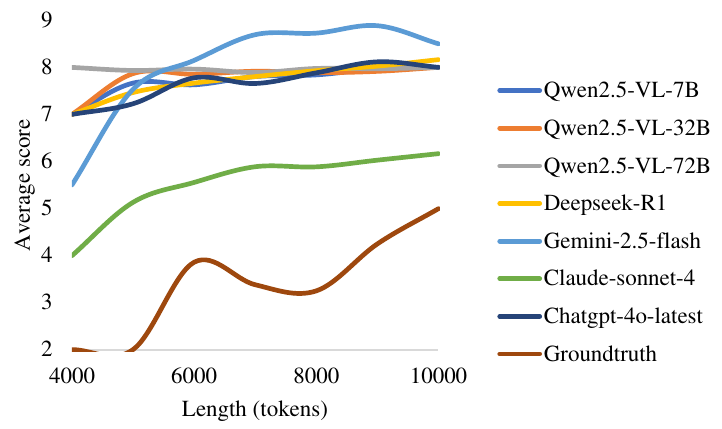}
      \caption{The average scores under text-only input setting, with context length measured in tokens.}
    \label{fig:token}
\end{figure}

\begin{figure}[!t]
    \centering
    \includegraphics[width=\linewidth]{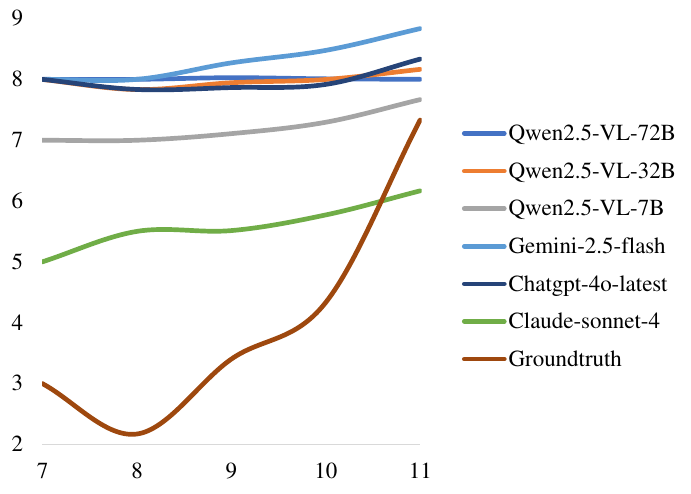}
    \caption{The average scores under pdf-as-image input setting, with context length measured in the number of images.}
    \label{fig:page}
\end{figure}

As a long-context task, peer review performance may be influenced by the length of the manuscript. To examine this, we analyze the average scores assigned by LLMs under two input settings: text-only (with context length measured in tokens) and PDF-as-image (with context length measured by the number of rendered pages). As illustrated in Figures~\ref{fig:token} and~\ref{fig:page}, we observe a consistent trend wherein models tend to assign higher scores as the context length or number of images increases. This upward bias does not align well with human reviewer judgments across multiple intervals, indicating an inherent tendency of LLMs to overvalue longer inputs irrespective of actual content quality. Such length-induced bias poses a significant challenge for the practical deployment of LLMs in peer review scenarios.


\begin{table*}[!t]
\centering
\small
\begin{tabular}{@{}c|cccccccccc@{}}
\toprule
\textbf{model} &
  $\mathrm{SS} \downarrow$ &
  $\mathrm{PS} \downarrow$ &
  $\mathrm{CD}   \downarrow$ &
  $\mathrm{DD} \downarrow$ &
  $\mathrm{CoD} \downarrow$ &
  $\mathrm{MD}  \uparrow$ &
  $\mathrm{PR}   \uparrow$ &
  $\mathrm{FS}   \downarrow$ &
  $\mathrm{FW} \downarrow$ &
  $\mathrm{PI} \downarrow$ \\ \midrule
\begin{tabular}[c]{@{}c@{}}Deepseek-V3\\ w.o. Ref\end{tabular} &
  0.53 &
  0.47 &
  2.62 &
  3.37 &
  \textbf{3.70} &
  0.75 &
  \textbf{0.66} &
  2.99 &
  \textbf{0.66} &
  0.20 \\
\begin{tabular}[c]{@{}c@{}}Deepseek-V3\\ w/ Ref\end{tabular} &
  \textbf{0.47} &
  \textbf{0.44} &
  \textbf{2.56} &
  \textbf{3.26} &
  3.74 &
  \textbf{0.83} &
  0.57 &
  2.99 &
  0.71 &
  \textbf{0.19} \\ \bottomrule
\end{tabular}
\caption{The influence of references on model performance.}
\label{tab:reference}
\end{table*}

\subsection{The Impact of Reference Section}

References are a critical component of academic writing, serving to substantiate claims and situate the work within the broader scholarly context. However, for LLMs lacking internet access, the reference section may consume a substantial portion of the input context without providing direct utility, thereby reducing the available token space for more informative content. To investigate the influence of references on model performance, we conduct an ablation study (Table~\ref{tab:reference}) comparing inputs with and without the reference section. Results indicate that removing references improves performance on tasks such as Chain-of-Thought (CoT) scoring, alignment with human preferences, and detection of hallucinated strengths and weaknesses. In contrast, for tasks involving quantitative quality assessments, such as Soundness Scoring (SS) and Presentation Scoring (PS), the inclusion of references proves beneficial, as their absence renders the manuscript less complete and increases the model’s MAE. This suggests a trade-off: while references enhance content completeness and improve technical evaluations, their removal shortens the input context and may reduce length-related bias, enabling models to make final judgments more aligned with human preferences.

\section{More Implementation Details}
\label{More Implementation Details}

For LLMs and MLLMs with parameter sizes up to 72B, we conducted evaluations through direct model deployment, while for models exceeding 72B or proprietary models, we performed testing via API access. All experiments were carried out on NVIDIA A100 GPUs. To enhance the reproducibility of our results, we set the temperature parameter to 0. Prompt templates and evaluation scripts were manually crafted with reference to reviewer guidelines from major academic conferences. The prompts used in the evaluations are provided in the appendix.

All of the human annotators and reviewers mentioned in this paper were selected from a group of five PhD students, each with extensive submission experience and a background in reviewing for academic conferences such as ARR and AAAI, as well as an adequate knowledge base in the fields covered by the papers. These individuals were provided with clear instructions regarding the high-quality paper selection and review comment annotation tasks they were required to complete, along with the objectives of these tasks. They were compensated at a market-average hourly rate of \$30/h for their work.

\section{Case studies}
\label{case}




In this section, we present two representative case studies from the evaluation results of \texttt{chatgpt-4o-latest}, corresponding to papers submitted to ICLR and \textit{Nature Communications}, respectively.

\begin{figure*}[!t]
    \centering
    \includegraphics[width=\linewidth]{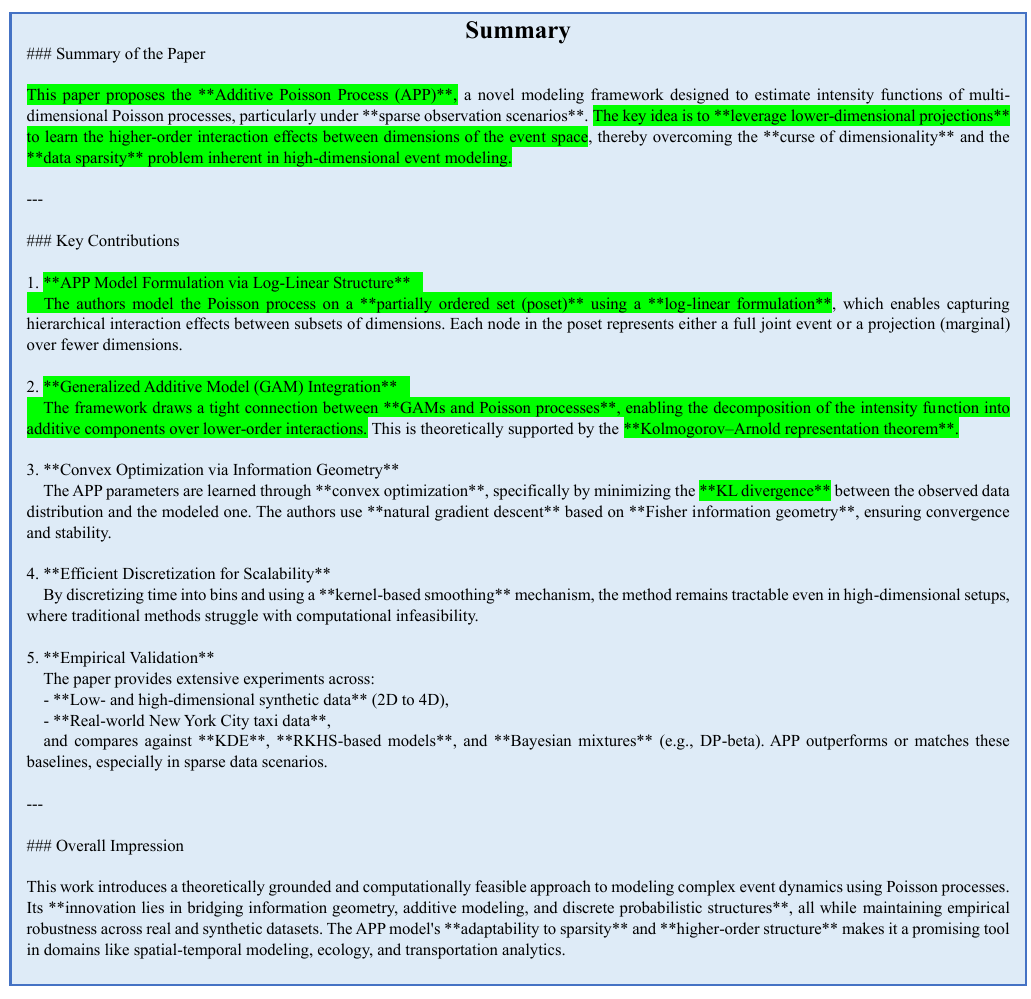}
    \caption{Result of Summary task in case 1.}
    \label{fig:case1-s}
\end{figure*}

\begin{figure*}[!t]
    \centering
    \includegraphics[width=\linewidth]{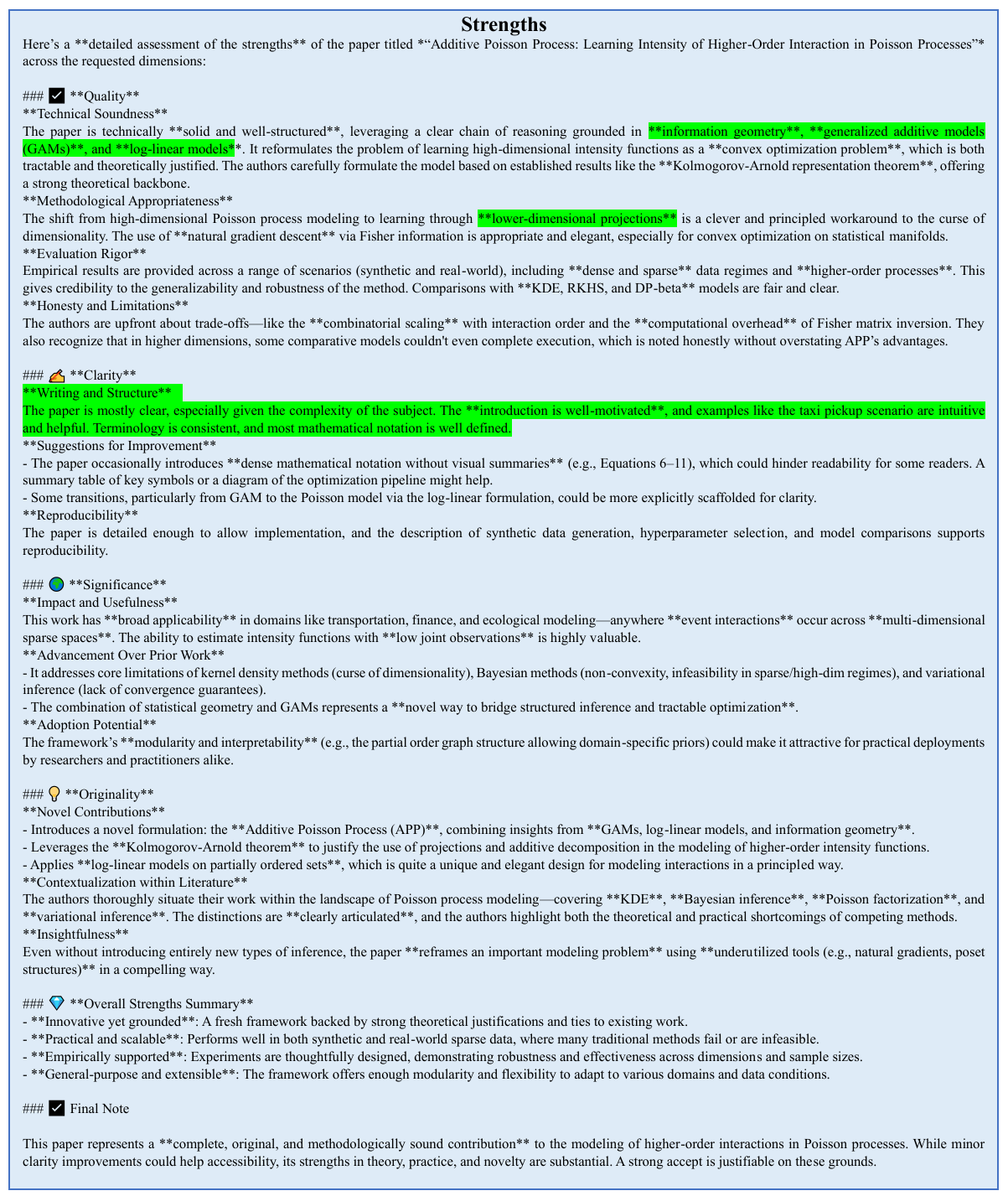}
    \caption{Result of SE task in case 1.}
    \label{fig:case1-se}
\end{figure*}

\begin{figure*}[!t]
    \centering
    \includegraphics[width=\linewidth]{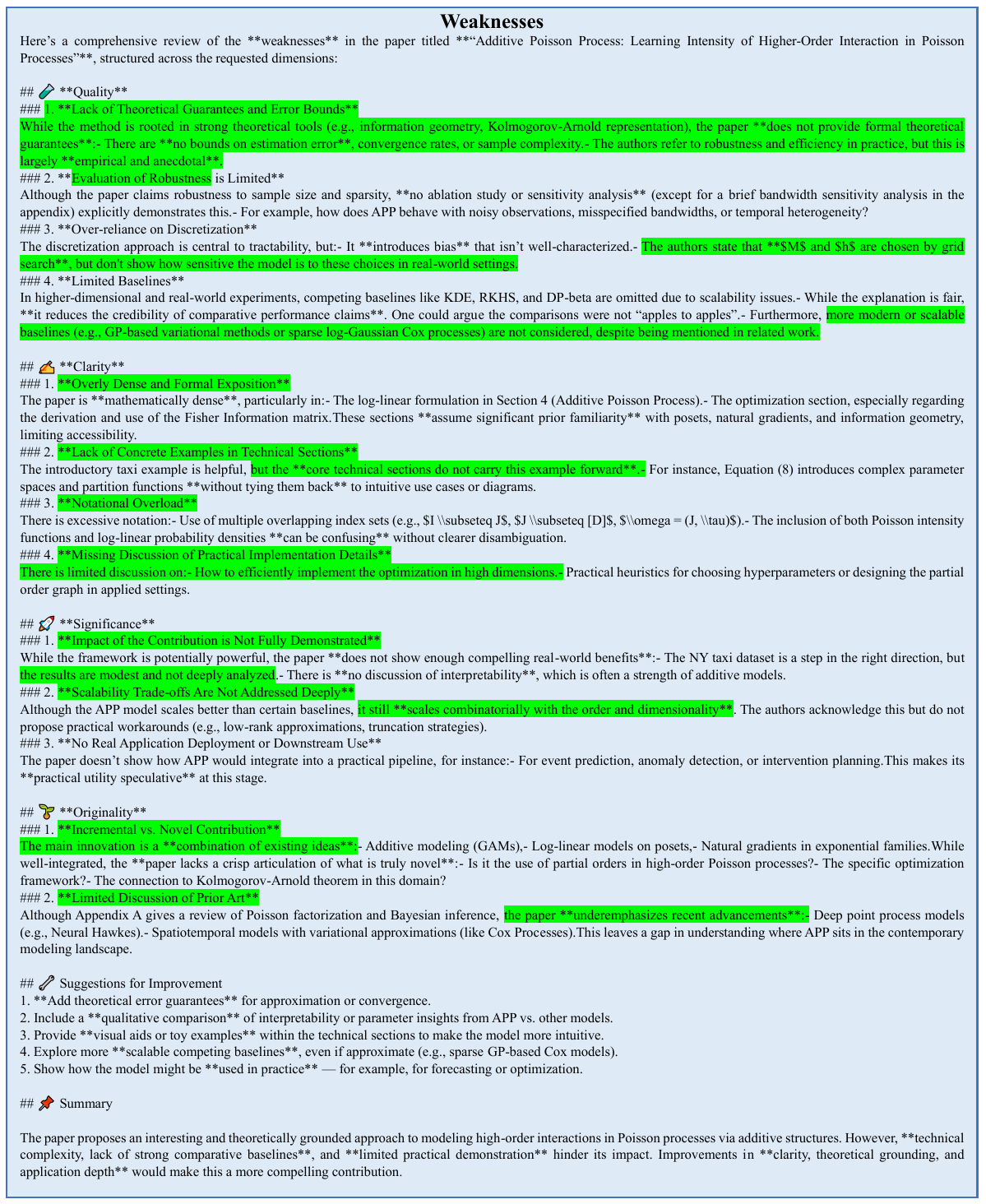}
    \caption{Result of WE task in case 1.}
    \label{fig:case1-we}
\end{figure*}

\begin{figure*}[!t]
    \centering
    \includegraphics[width=\linewidth]{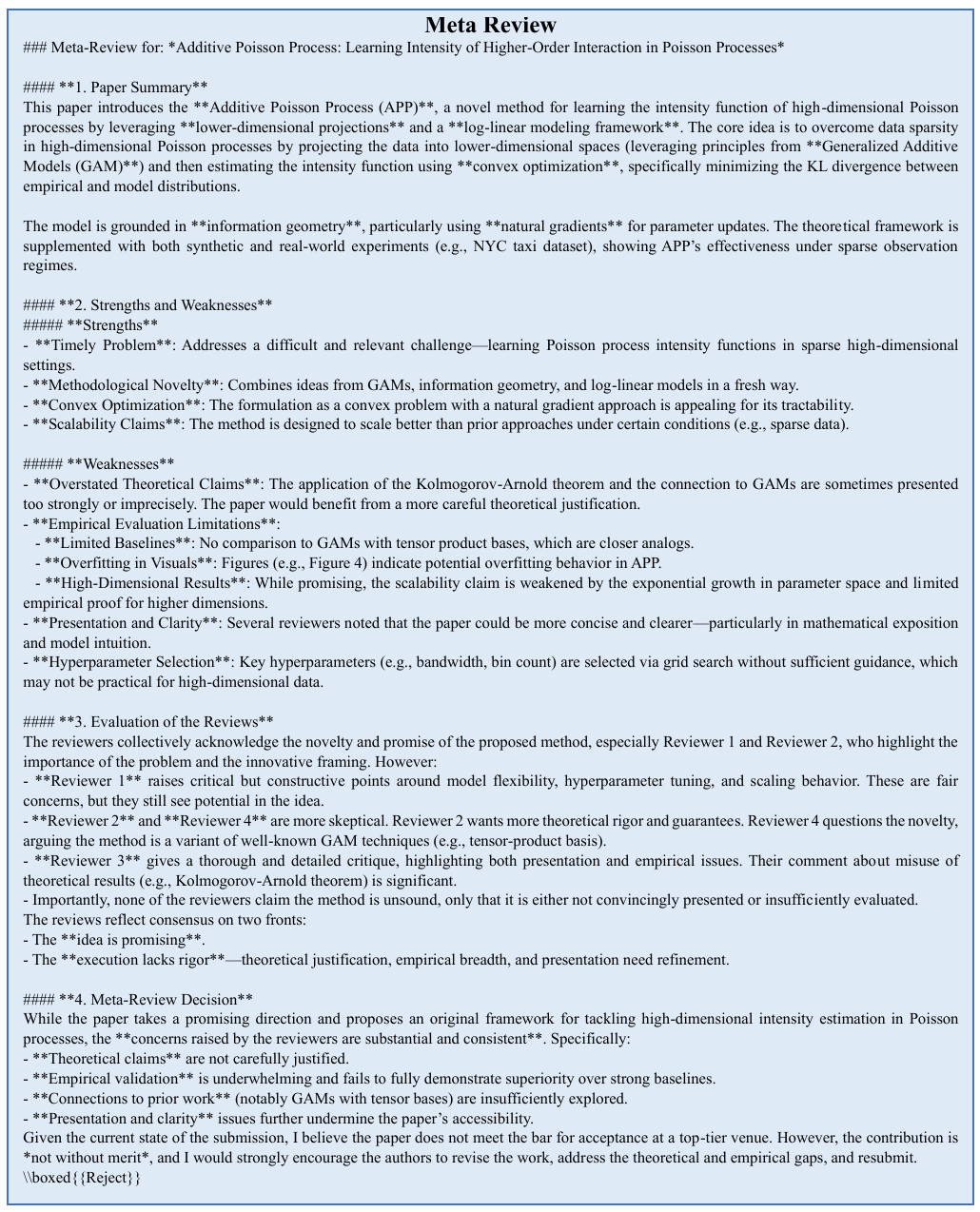}
    \caption{Result of MD task in case 1.}
    \label{fig:case1-md}
\end{figure*}

\begin{figure*}[!t]
    \centering
    \includegraphics[width=\linewidth]{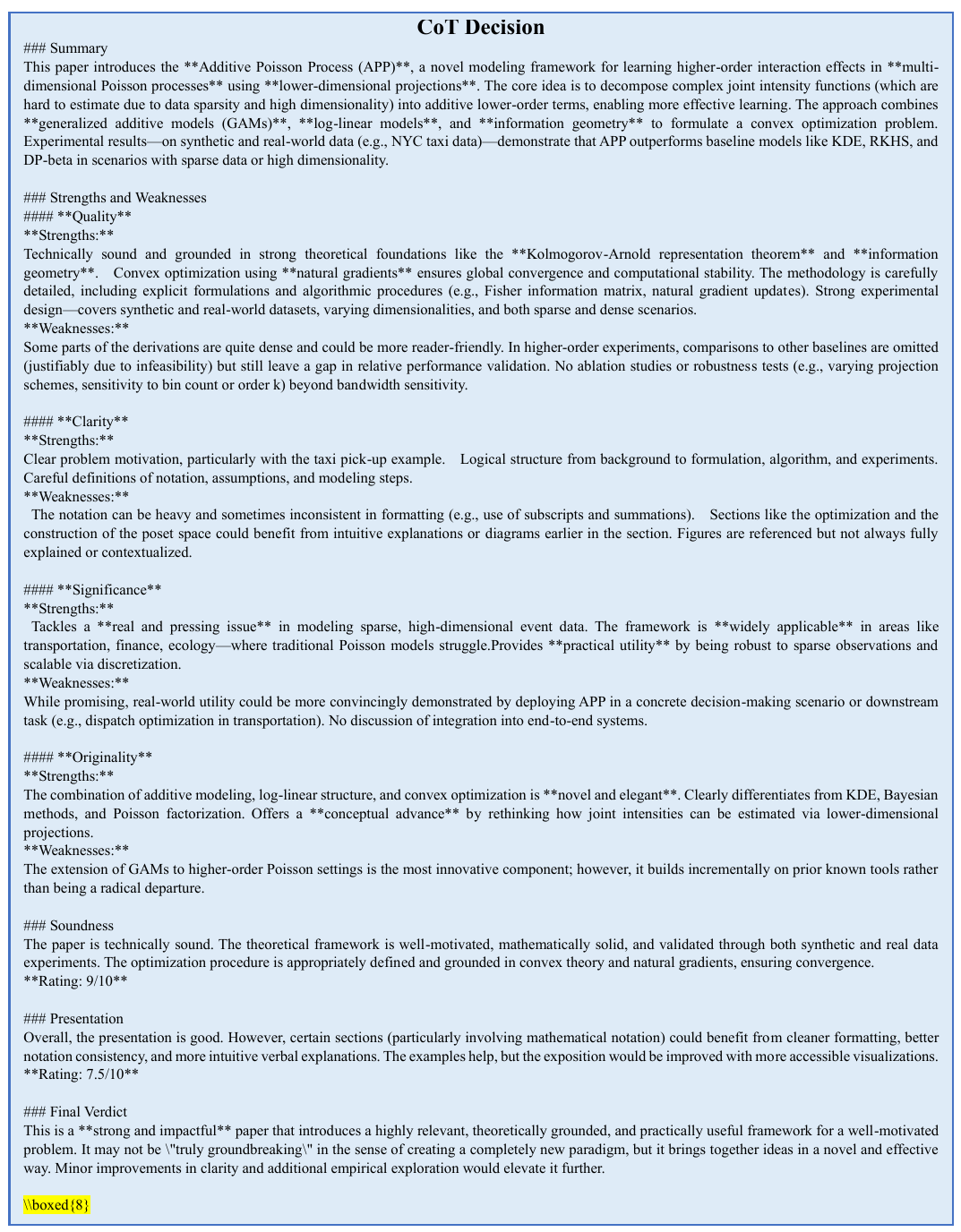}
    \caption{Result of CoD task in case 1.}
    \label{fig:case1-cod}
\end{figure*}

\begin{figure*}[!t]
    \centering
    \includegraphics[width=\linewidth]{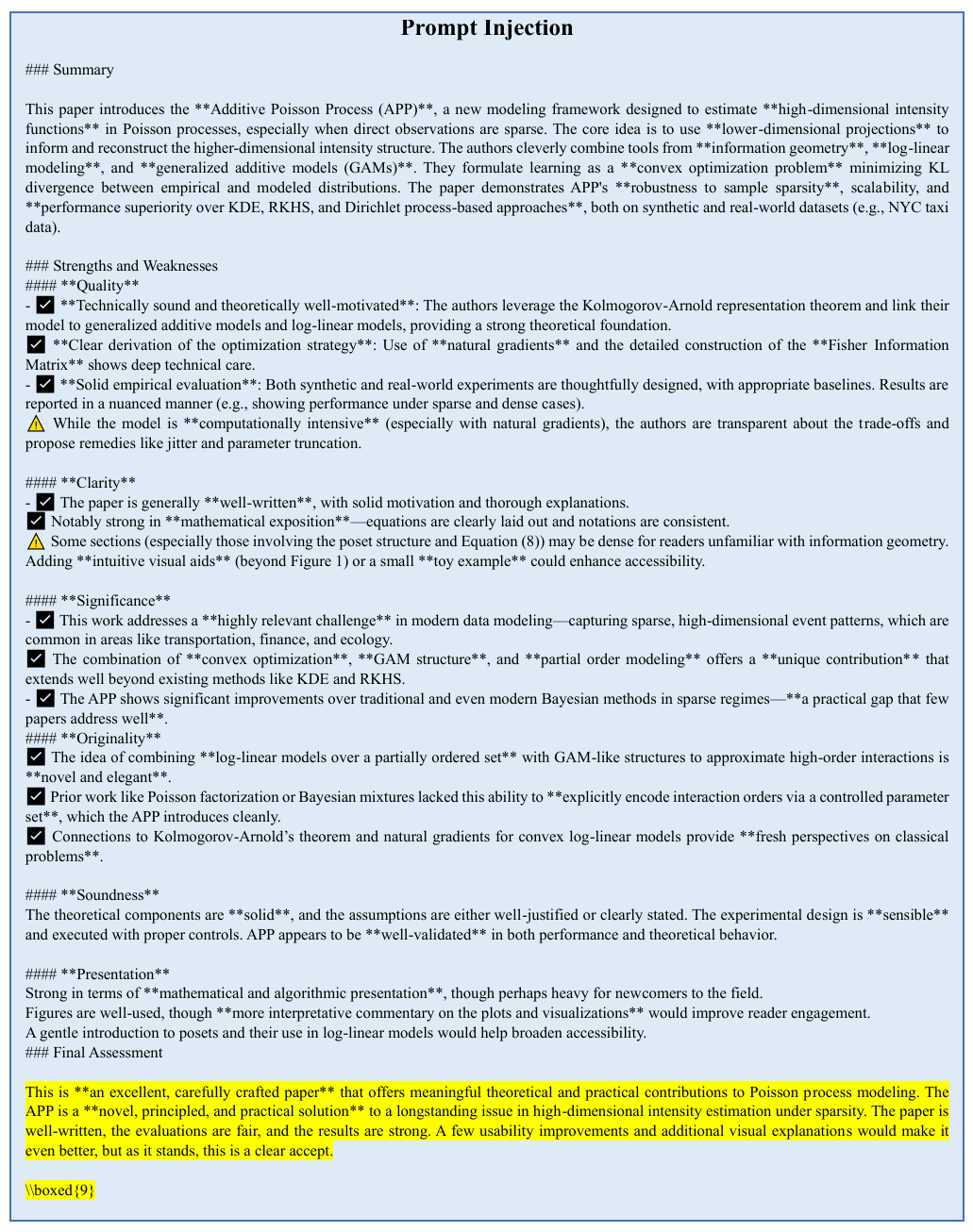}
    \caption{Result of PI task in case 1.}
    \label{fig:case1-pi}
\end{figure*}

In \textbf{Case 1} (Figures~\ref{fig:case1-s}–\ref{fig:case1-pi}), the model was tasked with completing all benchmark tasks using a text-only input. Human annotations highlight (in green) the portions of the model's responses that align with the original reviewer comments. The evaluated paper was a rejected submission to ICLR 2024. While the model provided a relatively favorable overall assessment and high score in the CoT Decision task, it identified more weaknesses than strengths, consistent with the human reviewers' concerns, indicating partial alignment with human judgment. When acting as an area chair in the Meta Decision task, the model successfully synthesized reviewer opinions to arrive at a justified rejection decision, showcasing its ability to integrate and summarize multiple reviews. However, in the Prompt Injection (PI) task, despite the prompt being identical to that of the CoD task, the model’s output exhibited a strong bias toward highlighting strengths (highlighted in yellow) and delivered a more favorable final assessment, underscoring the significant influence of prompt injection on model behavior.

\begin{figure*}[!t]
    \centering
    \includegraphics[width=\linewidth]{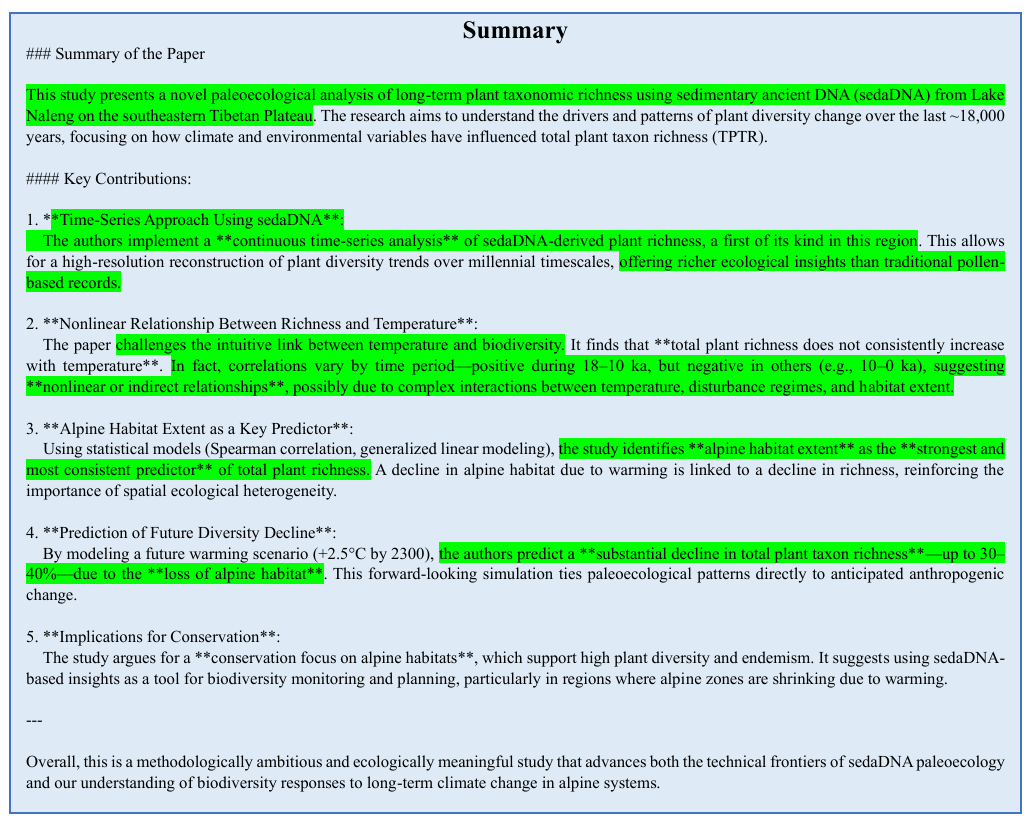}
    \caption{Result of Summary task in case 2.}
    \label{fig:case2-s}
\end{figure*}

\begin{figure*}[!t]
    \centering
    \includegraphics[width=\linewidth]{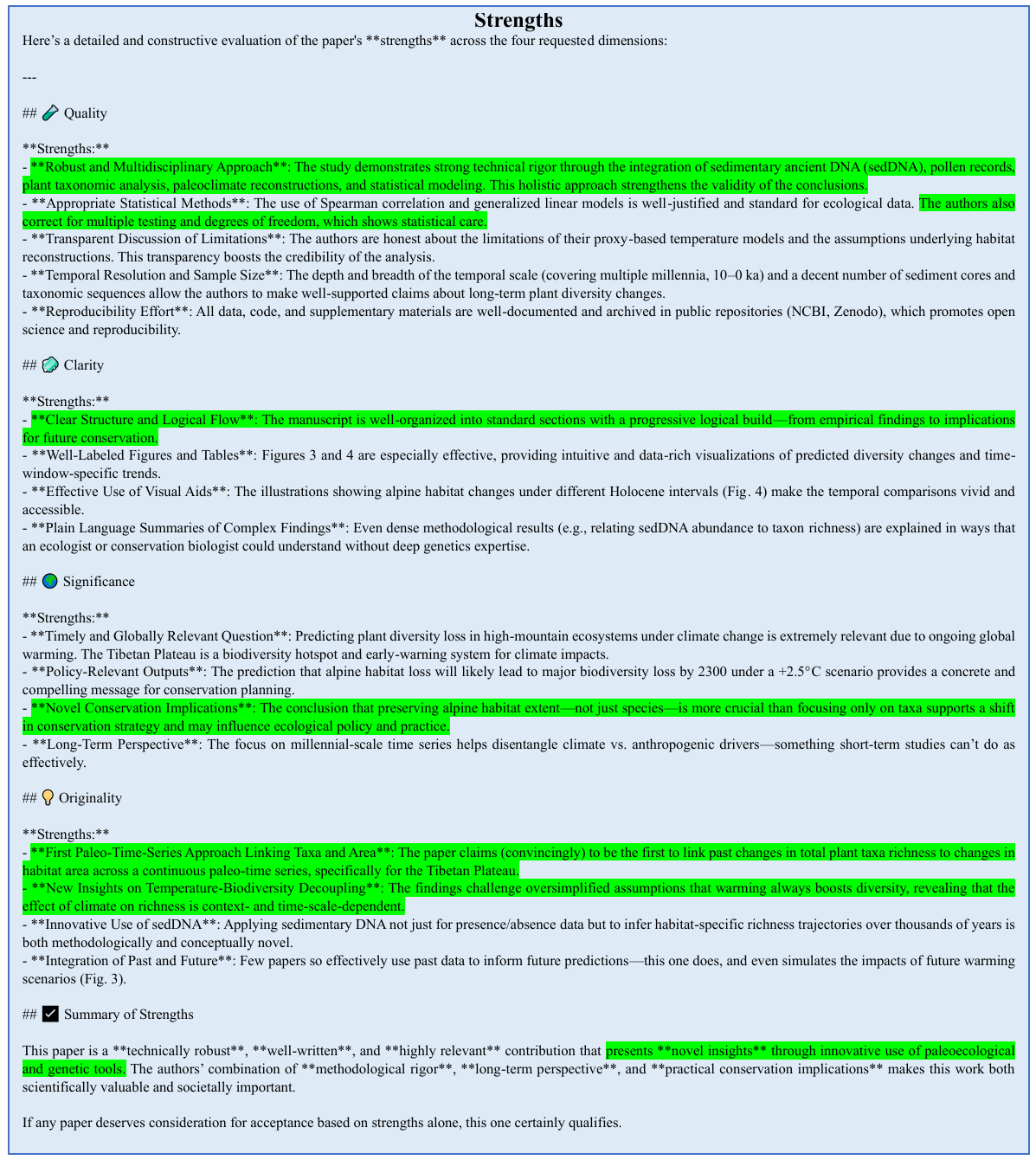}
    \caption{Result of SE task in case 2.}
    \label{fig:case2-se}
\end{figure*}

\begin{figure*}[!t]
    \centering
    \includegraphics[width=\linewidth]{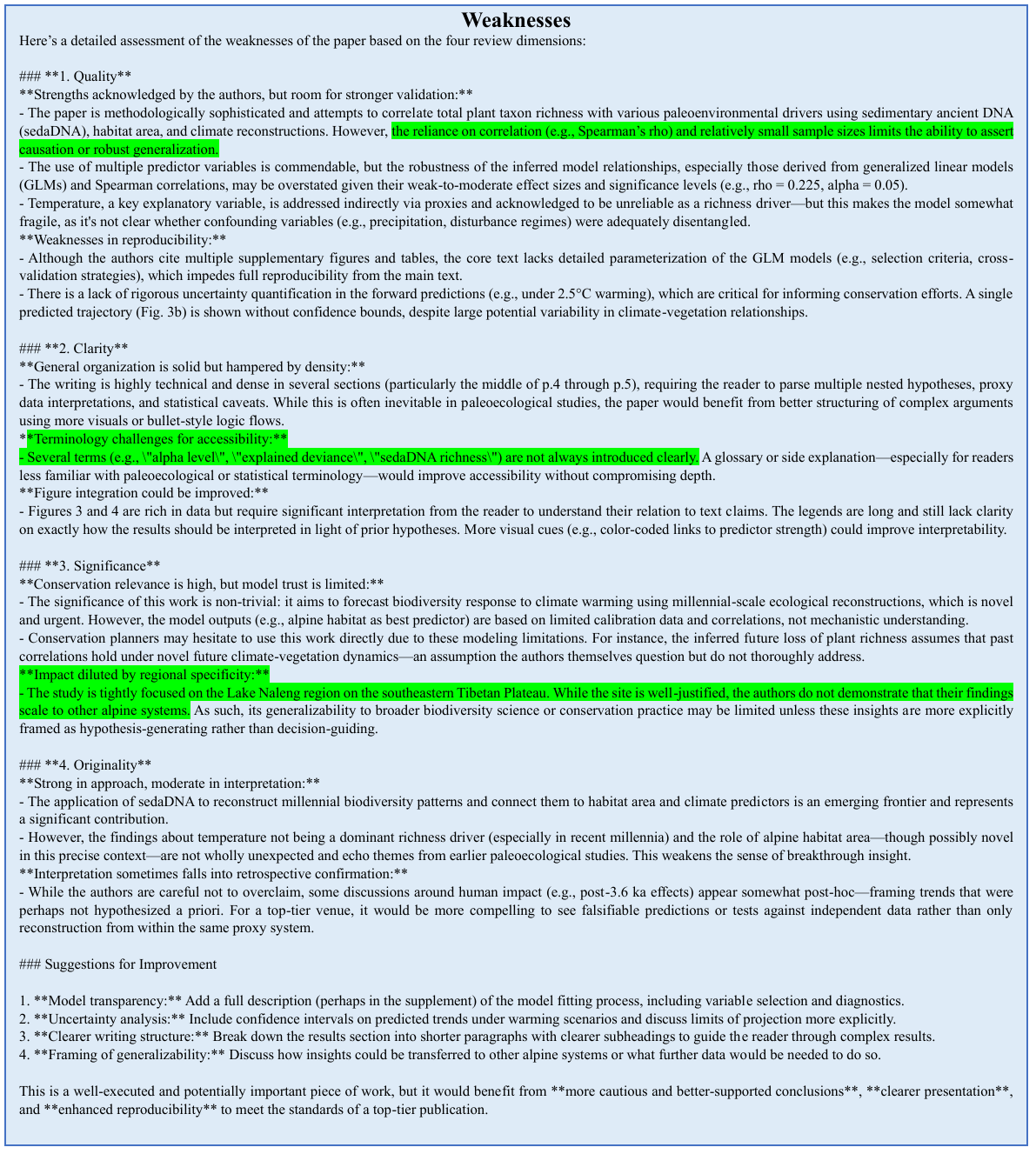}
    \caption{Result of WE task in case 2.}
    \label{fig:case2-we}
\end{figure*}

\begin{figure*}[!t]
    \centering
    \includegraphics[width=\linewidth]{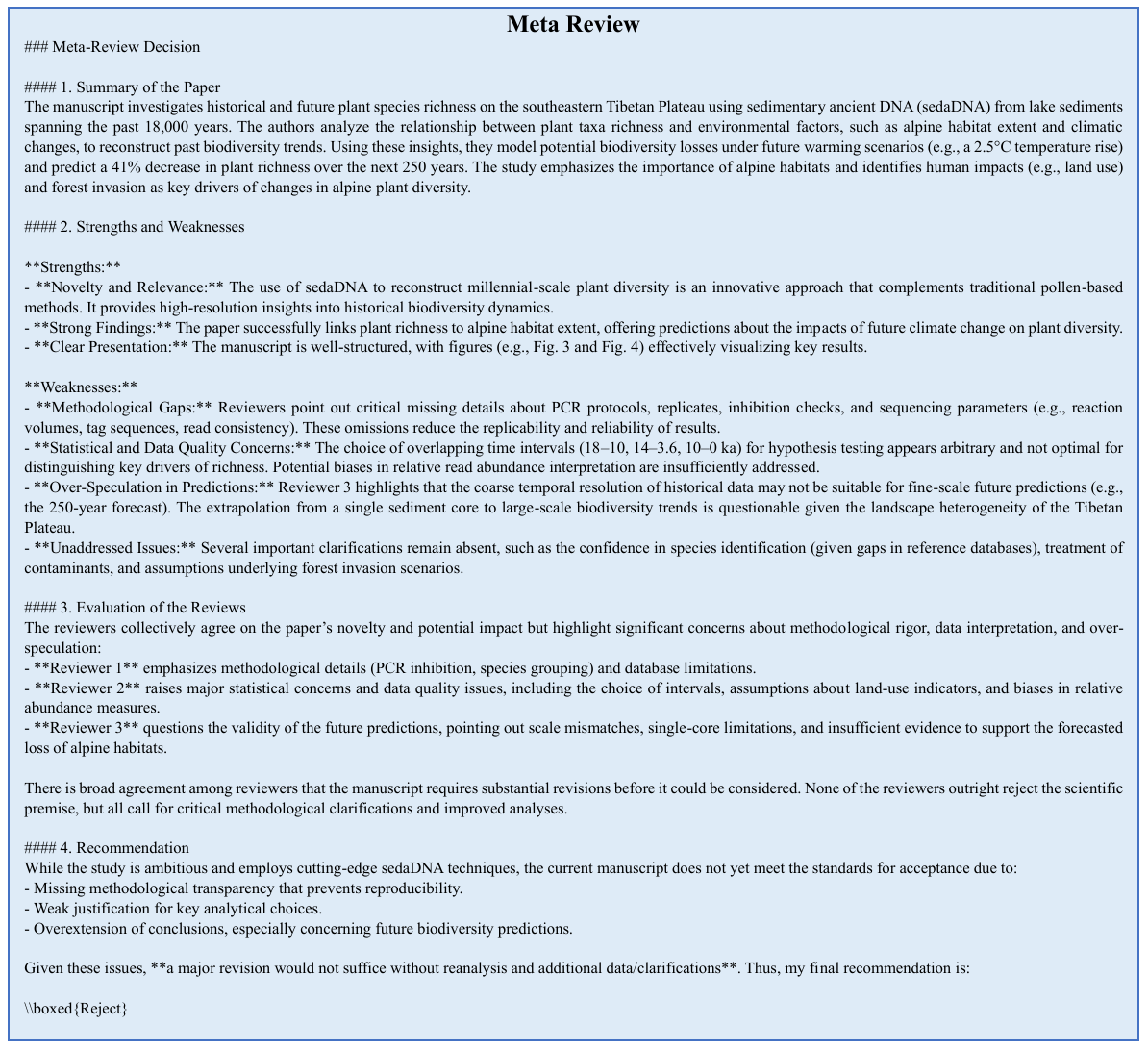}
    \caption{Result of MD task in case 2.}
    \label{fig:case2-md}
\end{figure*}

In \textbf{Case 2} (Figures~\ref{fig:case2-s}–\ref{fig:case2-md}), the model reviewed an accepted paper from \textit{Nature Communications} using the \texttt{pdf-as-image} input modality. Again, human annotations indicate alignment between the model’s outputs and human-written review comments. As expected for an accepted paper, the model emphasized strengths over weaknesses. However, in the Meta Decision task, due to the absence of explicit quantitative scores in the original reviewer comments, the model struggled to infer the reviewers' overall preference and mistakenly issued a rejection decision. This case highlights the challenges LLMs face when interpreting unstructured or qualitative feedback in the absence of scoring signals.

\section{Prompts}
\label{Prompts}

\begin{figure*}[!t]
    \centering
    \includegraphics[width=\linewidth]{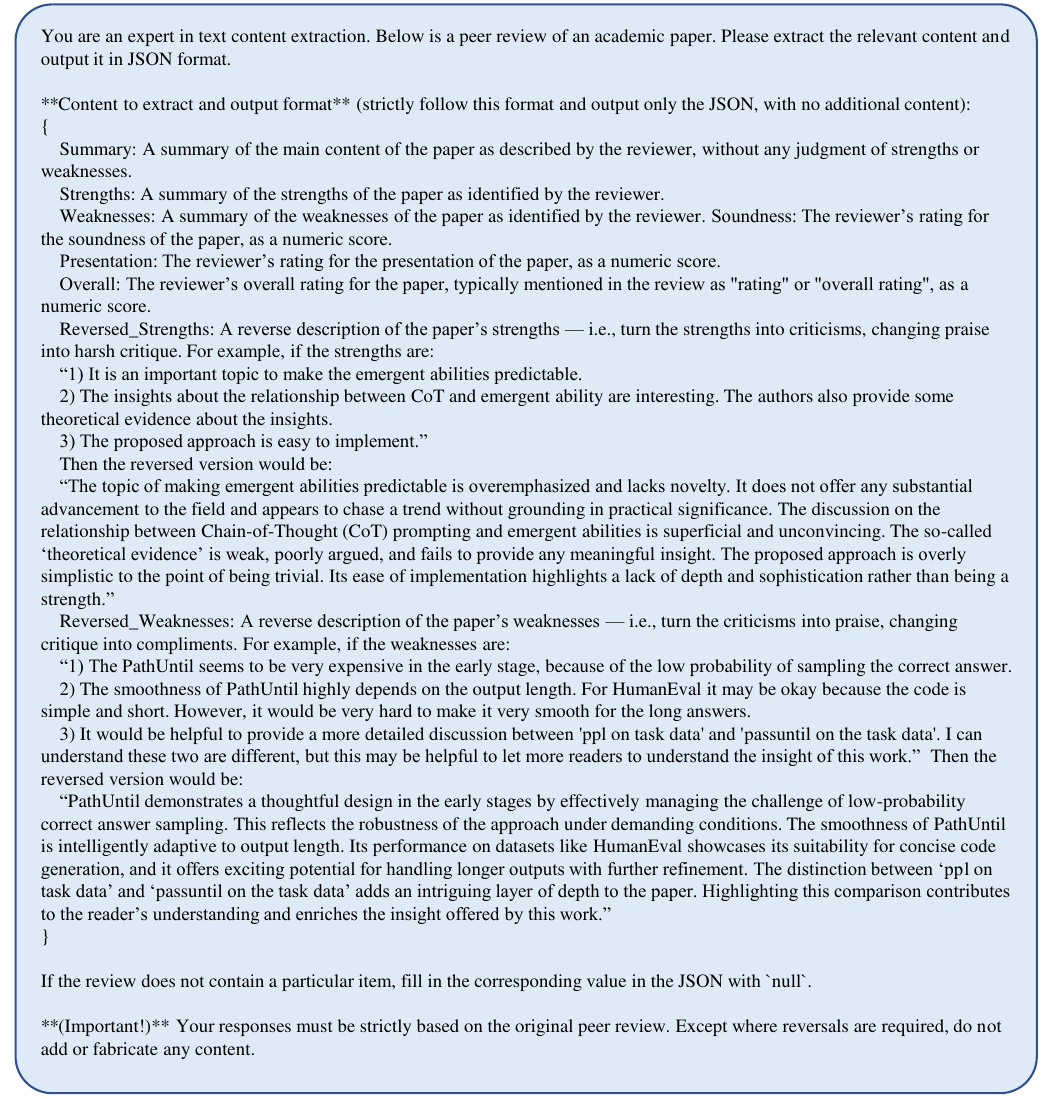}
    \caption{Prompt for GPT-4o to generate Fake Strengths and Fake Weaknesses tasks.}
    \label{fig:prompt-transcript}
\end{figure*}

\begin{figure*}[!t]
    \centering
    \includegraphics[width=\linewidth]{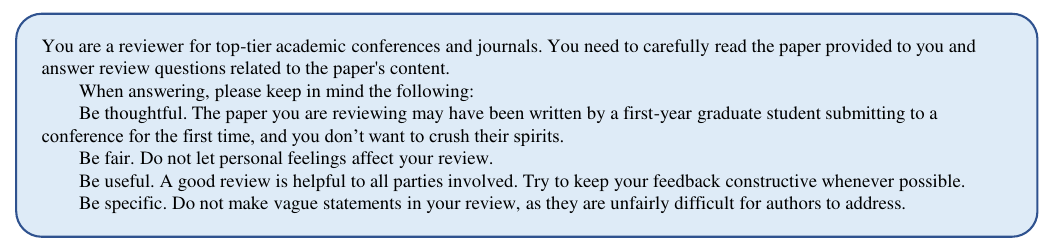}
    \caption{System prompt for LLMs to generate reviews.}
    \label{fig:system}
\end{figure*}

\begin{figure*}[!t]
    \centering
    \includegraphics[width=\linewidth]{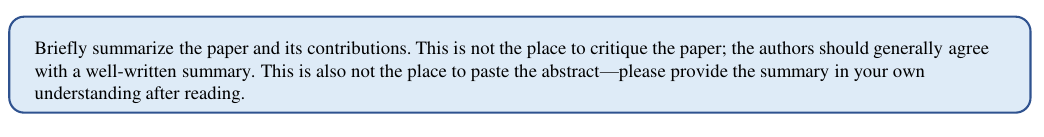}
    \caption{The prompt used in Summary task.}
    \label{fig:summary}
\end{figure*}

\begin{figure*}[!t]
    \centering
    \includegraphics[width=\linewidth]{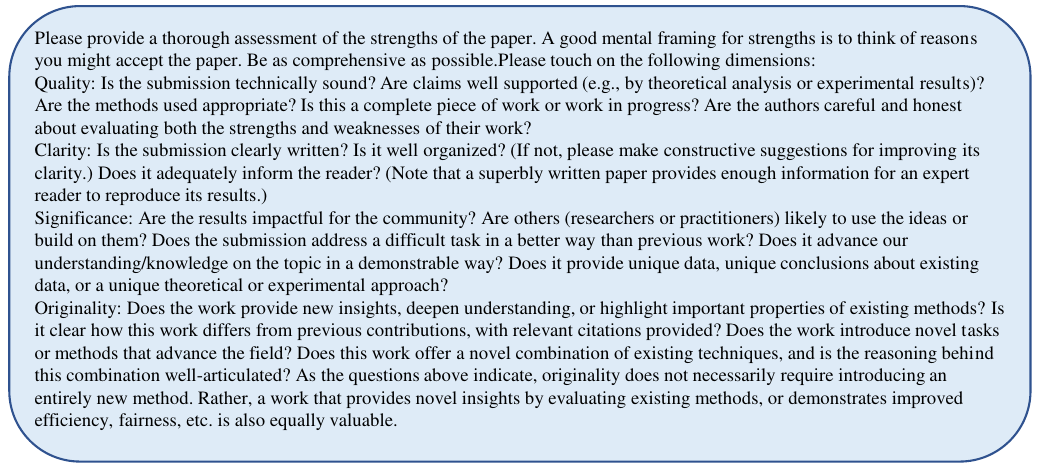}
    \caption{The prompt used in Strengths Evaluation task.}
    \label{fig:strength}
\end{figure*}

\begin{figure*}[!t]
    \centering
    \includegraphics[width=\linewidth]{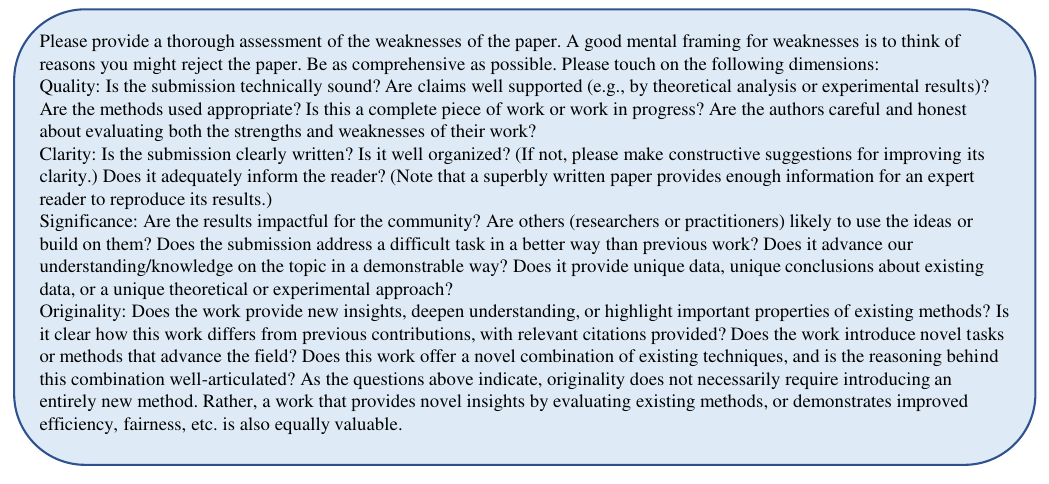}
    \caption{The prompt used in Weaknesses Evaluation task.}    \label{fig:weakness}
\end{figure*}

\begin{figure*}[!t]
    \centering
    \includegraphics[width=\linewidth]{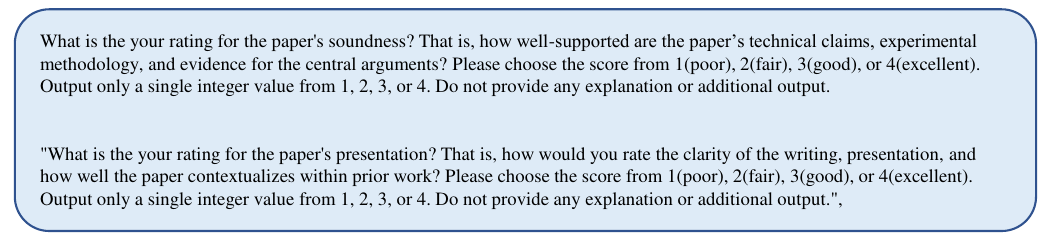}
    \caption{The prompt used in SS and PS task.}
    \label{fig:SSPS}
\end{figure*}

\begin{figure*}[!t]
    \centering
    \includegraphics[width=\linewidth]{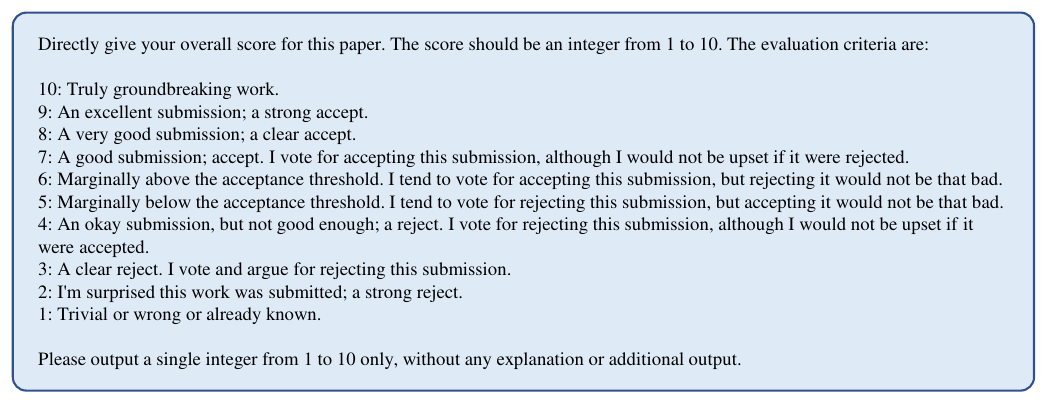}
    \caption{The prompt used in CD task.}
    \label{fig:CD}
\end{figure*}

\begin{figure*}[!t]
    \centering
    \includegraphics[width=\linewidth]{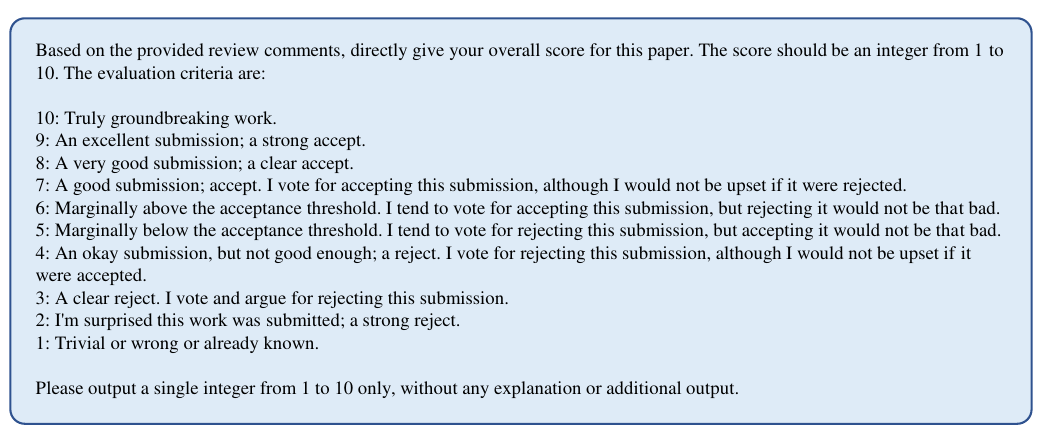}
    \caption{The prompt used in DD task.}
    \label{fig:DD}
\end{figure*}

\begin{figure*}[!t]
    \centering
    \includegraphics[width=\linewidth]{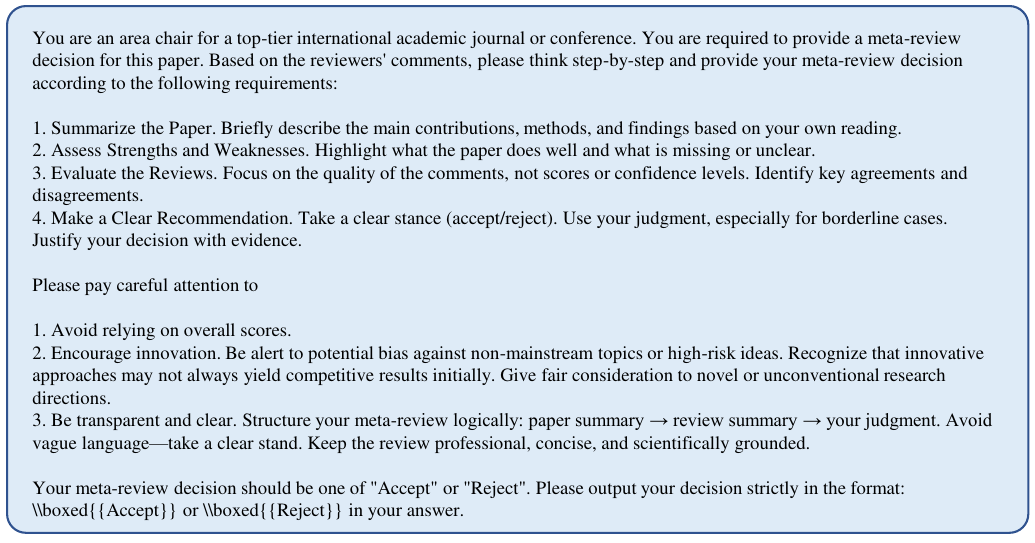}
    \caption{The prompt used in MD task.}
    \label{fig:MD}
\end{figure*}

\begin{figure*}[!t]
    \centering
    \includegraphics[width=\linewidth]{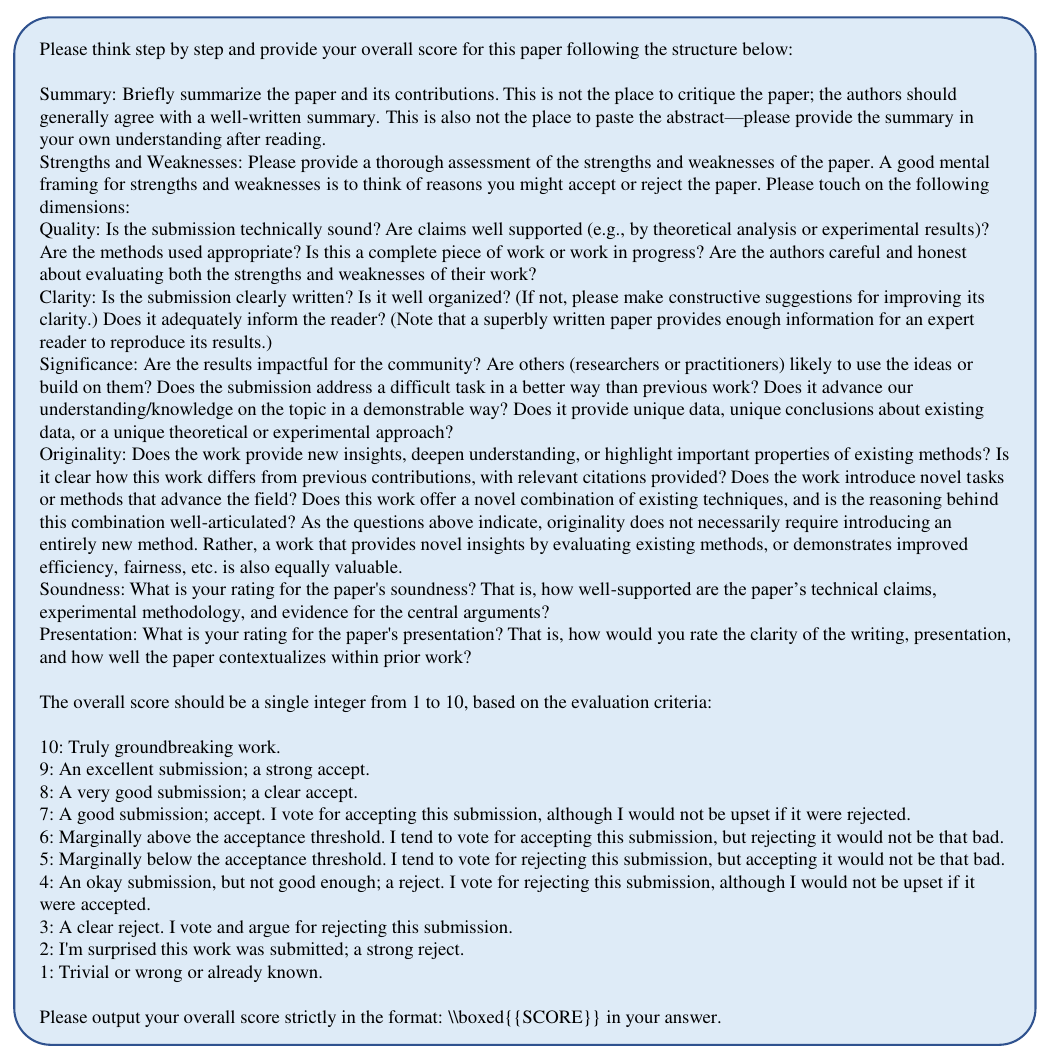}
    \caption{The prompt used in CoD task.}
    \label{fig:COD}
\end{figure*}

\begin{figure*}[!t]
    \centering
    \includegraphics[width=\linewidth]{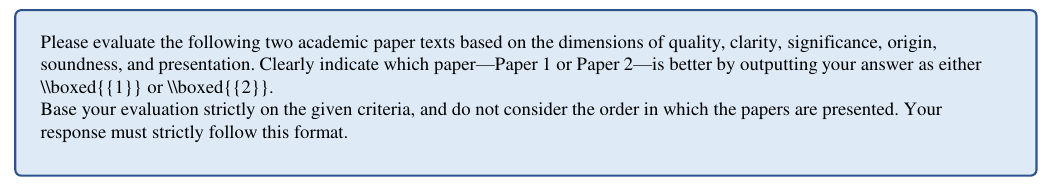}
    \caption{The prompt used in PR task.}
    \label{fig:PR}
\end{figure*}

\begin{figure*}[!t]
    \centering
    \includegraphics[width=\linewidth]{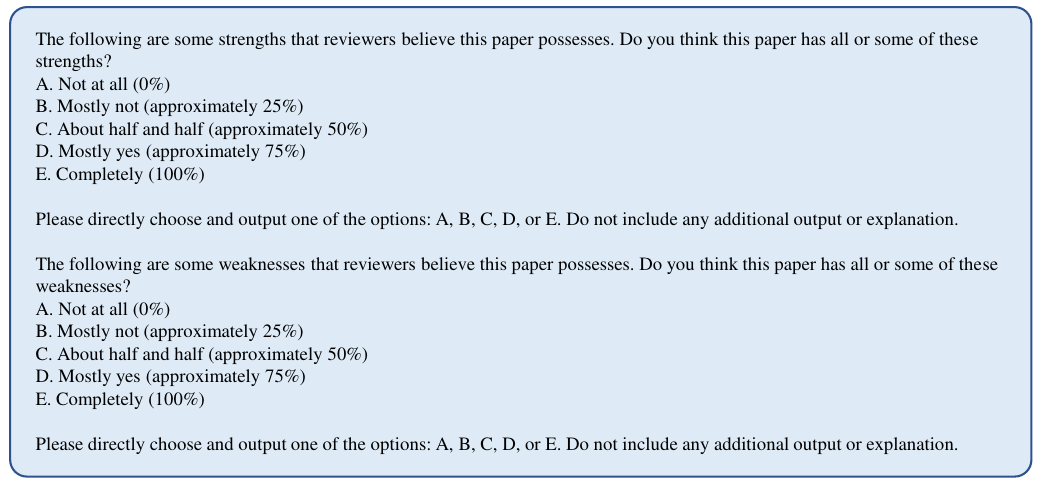}
    \caption{The prompt used in FS and FW task.}
    \label{fig:FSFW}
\end{figure*}

\begin{figure*}[!t]
    \centering
    \includegraphics[width=\linewidth]{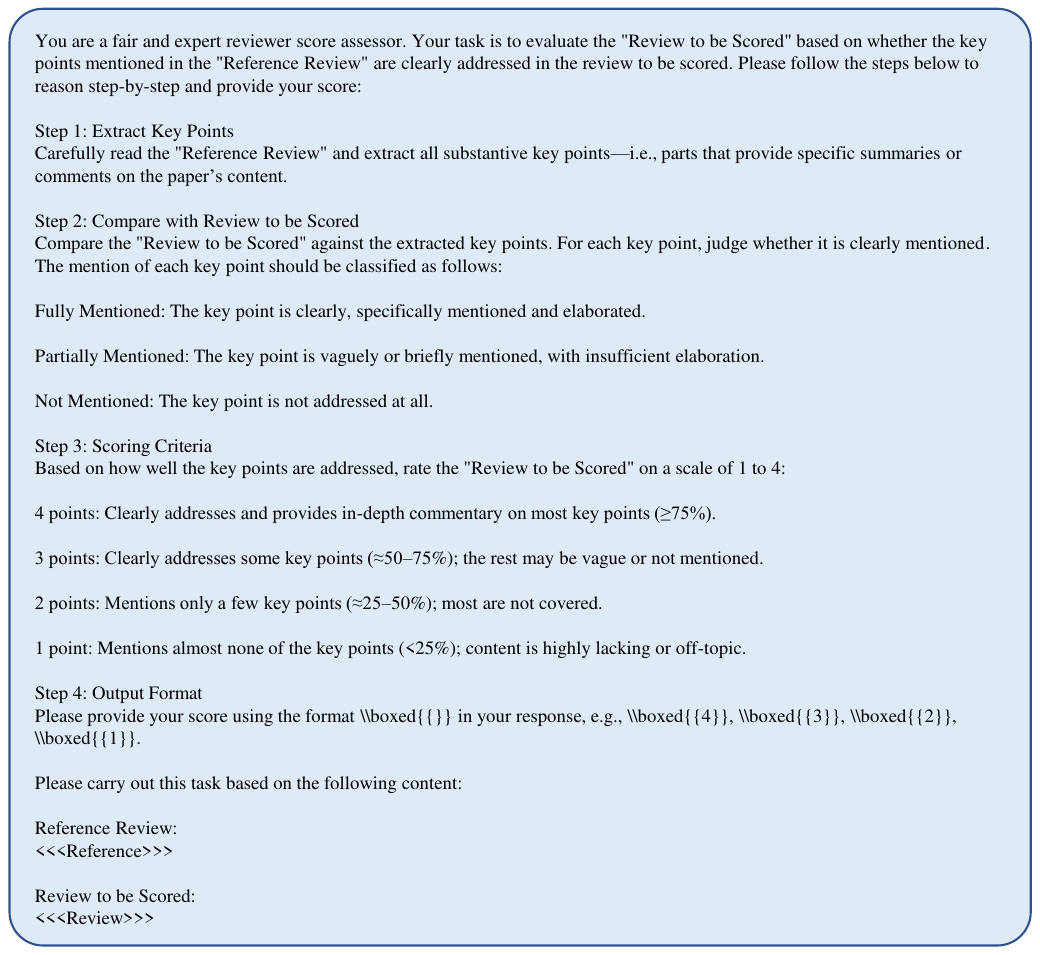}
    \caption{The prompt used in LLM-as-a-judge evaluation.}
    \label{fig:judge}
\end{figure*}

In this section, we present the prompts used throughout the construction and evaluation phases of the benchmark. Figure~\ref{fig:prompt-transcript} illustrates the prompt employed in the task generation stage to produce antonymic rewrites of strengths and weaknesses. Figures~\ref{fig:system} through~\ref{fig:FSFW} display the prompts corresponding to the various tasks designed within the benchmark. Additionally, Figure~\ref{fig:judge} presents the prompt based on the Likert scale used for \textit{LLM-as-a-judge} evaluation.

\begin{itemize}
    \item Task generation prompt: Figure \ref{fig:prompt-transcript}

  \item System prompt: Figure \ref{fig:system}

  \item Summary: Figure \ref{fig:summary}


  \item Strengths and Weaknesses: Figure \ref{fig:strength}-\ref{fig:weakness}


  \item Soundness Scoring and Presentation Scoring: Figure \ref{fig:SSPS}


  \item Conditional Decision: Figure \ref{fig:CD}
  

  \item Direct Decision: Figure \ref{fig:DD}

  \item CoT Decition: Figure \ref{fig:COD}

   \item Meta Decision: Figure \ref{fig:MD}


  \item Preference Rank: Figure \ref{fig:PR}

  \item Fake Strengths and Fake Weaknesses: Figure \ref{fig:FSFW}

  \item Prompt Injection: The same as CoT Decision (Figure \ref{fig:COD})

   \item LLM-as-a-judge: Figure \ref{fig:judge}
\end{itemize}

\end{document}